\documentclass[10pt,journal,compsoc]{IEEEtran}

\makeatletter
\def\endthebibliography{%
	\def\@noitemerr{\@latex@warning{Empty `thebibliography' environment}}%
	\endlist
}
\makeatother

\ifCLASSOPTIONcompsoc
  \usepackage[nocompress]{cite}
\else
  \usepackage{cite}
\fi

\usepackage{hyperref}       
\hypersetup{colorlinks=true,urlcolor=black}
\usepackage{url}            
\usepackage{booktabs}       
\usepackage{amsfonts}       
\usepackage{nicefrac}       
\usepackage{microtype}      
\usepackage{algorithm}
\usepackage{algorithmic}
\usepackage{graphicx}
\usepackage{epsfig}
\usepackage{amsmath}
\usepackage{amssymb}
\usepackage{amsthm}
\usepackage{subfigure}
\usepackage{booktabs} 
\usepackage{multirow}
\usepackage{etoolbox}
\usepackage{sail}
\usepackage{color}
\usepackage{footnote}
\usepackage{arydshln}
\usepackage{enumitem}
\usepackage{times}
\usepackage{ragged2e}
\usepackage{threeparttable}
\usepackage[export]{adjustbox}
\newcommand{\jing}[1]{{\color{black}#1}}
\newcommand{\guo}[1]{{\color{black}#1}}%
\newcommand{\liu}[1]{{\color{black}#1}}
\newcommand{\ice}[1]{{\color{black}#1}}
\newcommand{\yong}[1]{{\color{black}#1}}%

\newcommand{\tabincell}[2]{\begin{tabular}{@{}#1@{}}#2\end{tabular}}

\begin{document}

\title{Discrimination-aware Network Pruning \\ for Deep Model Compression}

\author{
	Jing Liu$^*$, Bohan Zhuang$^*$, Zhuangwei Zhuang$^*$, Yong Guo, Junzhou Huang, Jinhui Zhu, Mingkui Tan$^*$$^\dagger$
	\IEEEcompsocitemizethanks{
	    \IEEEcompsocthanksitem Jing Liu is with the School of Software Engineering, South China University of Technology and also with the Key Laboratory of Big Data and Intelligent Robot (South China University of Technology), Ministry of Education. E-mail: seliujing@mail.scut.edu.cn
		\IEEEcompsocthanksitem Zhuangwei Zhuang, Yong Guo, Jinhui Zhu, and Mingkui Tan are with the School of Software Engineering, South China University of Technology. Mingkui Tan is also with the Pazhou Laboratory, Guangzhou, China. E-mail: \{z.zhuangwei, guo.yong\}@mail.scut.edu.cn, \{csjhzhu, mingkuitan\}@scut.edu.cn.
		\IEEEcompsocthanksitem Bohan Zhuang is with the Faculty of Information Technology, Monash University, Australia. E-mail: bohan.zhuang@monash.edu.
		\IEEEcompsocthanksitem Junzhou Huang is with Tencent AI Lab, Shenzhen, Guangdong, China. E-mail: joehhuang@tencent.com.
		\IEEEcompsocthanksitem $^*$ Authors contributed equally. $^\dagger$ Corresponding author.
	}
}

\IEEEtitleabstractindextext{%
\begin{abstract}
\justifying
\jing{
We study network pruning which aims to remove redundant channels/kernels and hence speed up the inference of deep networks.
}
Existing pruning methods either train from scratch with sparsity constraints or minimize the reconstruction error between the feature maps of the pre-trained models and the compressed ones. Both strategies suffer from some limitations: the former kind is computationally expensive and difficult to converge, while the latter kind optimizes the reconstruction error but ignores the discriminative power of channels. In this paper, we \guo{propose} a simple-yet-effective method called discrimination-aware channel pruning (DCP) to choose the channels that actually contribute to the discriminative power. To this end, we first introduce additional discrimination-aware losses into the network to increase the discriminative power of the intermediate layers. \jing{Next}, we select the most discriminative channels for each layer \jing{by considering the discrimination-aware loss and the reconstruction error, simultaneously. We then formulate channel pruning as a sparsity-inducing optimization problem with a convex objective and propose a greedy algorithm to solve the resultant problem.
Note that a channel (3D tensor) often consists of a set of kernels (each with a 2D matrix). 
\guo{Besides the redundancy in channels}, some kernels in a channel may also \guo{be redundant and fail to} contribute to the discriminative power of the network, resulting in kernel level redundancy.
To solve this {issue}, we  propose a discrimination-aware kernel pruning (DKP) method to \guo{further compress deep networks} by removing redundant kernels.
{To avoid manually determining the pruning rate for each layer,}
we propose {two} adaptive stopping conditions to automatically {determine the} number of selected channels/kernels. {The proposed adaptive stopping conditions tend to yield} 
more {efficient} models with better performance {in practice}. 
Extensive experiments on both image classification and face recognition demonstrate the effectiveness of our methods. For example, on ILSVRC-12, \guo{the resultant} ResNet-50 \guo{model} with 30\% reduction of channels even outperforms the baseline model by 0.36\% \guo{in terms of} Top-1 accuracy. We also deploy the pruned models on a smartphone (equipped with a Qualcomm Snapdragon 845 processor). The pruned MobileNetV1 and MobileNetV2 achieve 1.93$\times$ and 1.42$\times$ inference acceleration on the mobile device, respectively, with negligible performance degradation. 
\guo{The source code and the pre-trained models are available at \url{https://github.com/SCUT-AILab/DCP}.}
}
\end{abstract}

\begin{IEEEkeywords}
\jing{Channel Pruning, Kernel Pruning, Network Compression, Deep Neural Networks.}
\end{IEEEkeywords}}

\maketitle

\IEEEdisplaynontitleabstractindextext

%
\IEEEpeerreviewmaketitle

\IEEEraisesectionheading{\section{Introduction}\label{sec:introduction}}
\IEEEPARstart{S}{ince} 2012, deep neural networks (DNNs) have achieved great success in many computer vision tasks,~\eg, image classification~\cite{krizhevsky2012imagenet,srivastava2015training,he2016deep}, face recognition~\cite{Schroff_2015_CVPR,sun2015deepid3,Deng2019}, object detection~\cite{redmon2016you,ren2016faster,liu2016ssd}, image generation~\cite{goodfellow2014generative,pmlr-v80-cao18a,guo2019auto} and video analysis~\cite{simonyan2014two,wang2016temporal,zeng2019breaking}. 
However, the large model size and high computational costs remain
great obstacles for many applications, especially on some constrained devices with
limited memory and computational resources.
To address this problem, model compression is an effective approach, which aims to reduce the model redundancy without significant degeneration in performance.

Recent studies on model compression mainly contain three categories: quantization~\cite{han2015deep,rastegari2016xnor,zhou2017}, sparse or low-rank compression~\cite{guo2016dynamic,han2015deep,zhang2016accelerating}, and network pruning~\cite{liu2017learning,luo2018thinet,yu2017nisp,ye2018rethinking}. 
Network quantization seeks to represent weights and activations with low bitwidth fixed-point integers, thus convolution
operations can be implemented by efficient XNOR-popcount bitwise operations for substantial
speedup. However, the training can be very difficult \guo{since the non-differentiable quantizer that transforms the continuous weights/activations into the discrete ones would inevitably bring errors and hamper the model performance~\cite{rastegari2016xnor}.}
Sparse connections methods can \guo{obtain} a high compression rate in theory, but they may generate irregular convolutional kernels that need \guo{carefully designed} sparse matrix operations. 
\jing{Low-rank compression methods \guo{seek to} approximate the \guo{original filters of a} pre-trained model with low-rank filters. Nevertheless, they are often inefficient for \guo{the convolutions with small kernels sizes, \eg, $1 {\times} 1$~\cite{TaiXWE15}}.
}  
In contrast, network pruning reduces the model size and speeds up the inference by removing the redundant modules (channels~\cite{he2017channel, zhuang2018discrimination, liu2017learning} or kernels~\cite{anwar2017structured, molchanov2016pruning}). In particular, channel pruning can be well supported by existing deep learning libraries with little additional effort compared with network quantization and sparse or low-rank connections.
\guo{More critically, most compression methods, such as quantization, can be easily applied on top of network pruning. For example, pruning redundant channels/kernels is able to further \liu{reduce the model size and} accelerate the inference speed of the quantized models by reducing the number of parameters~\cite{han2015deep}.
}

\jing{In network pruning,}
how to identify the informative (or important) channels/kernels (also known as channel/kernel selection) 
\liu{is an important problem.}
\jing{Existing methods 
\guo{can be divided into two categories},
namely, training-from-scratch methods~\cite{alvarez2016learning,liu2017learning,wen2016learning} and reconstruction-based methods~\cite{he2017channel,hu2016network,li2016pruning,luo2018thinet}.
Training-from-scratch methods directly learn the importance of channels/kernels with sparsity regularization, but it is very difficult to train very deep networks on large-scale datasets~\cite{alvarez2016learning,liu2017learning}}. The reconstruction-based methods seek to perform network pruning by minimizing the reconstruction error of feature maps between the pruned model and the pre-trained one~\cite{he2017channel,luo2018thinet}. {However, the performance is highly affected by the quality of the pre-trained model. If the pre-trained model is not well trained, the pruning performance can be limited. More importantly, }
these methods 
suffer from a critical limitation: the redundant channels/kernels may be mistakenly kept to minimize the reconstruction error of feature maps. Consequently, these methods may 
\guo{incur severe performance degradation}
on more compact and deeper models, such as MobileNet~\cite{howard2017mobilenets,sandler2018inverted} for large-scale datasets.
	
In this paper, we aim to overcome the drawbacks of both strategies. In contrast to existing methods~\cite{he2017channel,hu2016network,li2016pruning,luo2018thinet}, we assume and highlight that an informative channel/kernel, no matter where it is, should have \guo{sufficient} discriminative power; otherwise, it should be removed. \jing{Based on this intuition, we propose a discrimination-aware channel pruning (DCP) method to find the channels that actually contribute to the discriminative power of the network. \guo{In DCP}, relying on a pre-trained model, we first introduce multiple additional discrimination-aware losses into the network to increase the discriminative power of the intermediate layers. 
Then, we perform channel selection to find the most discriminative channels for each layer by considering both the discrimination-aware loss and the reconstruction error of feature maps.} 
In this way, we are able to make a balance between the discriminative power of the channels and feature maps reconstruction. 
\jing{
Note that a channel (3D tensor) \ice{consists} of \liu{a set of} kernels (each with a 2D matrix). In practice, some kernels in the selected channels may be redundant and fail to contribute to the discriminative power of the network, which leads to kernel level redundancy.
To solve this issue, we propose a discrimination-aware kernel pruning (DKP) method to find the kernels with discriminative power.
}

Our main contributions are summarized as follows. 
\begin{itemize}
\item
We propose a discrimination-aware channel/kernel pruning (DCP/DKP) scheme to compress deep models with the introduction of additional discrimination-aware losses. \jing{The proposed methods first fine-tune the model with the additional losses and the final objective}. 
\jing{Then, we conduct channel/kernel selection by simultaneously considering the additional loss and the reconstruction error of feature maps. In this way, the proposed method is able to find the channels/kernels that \liu{actually} contribute to the discriminative power of the network.}

\item
We formulate the channel/kernel selection problem as a constrained optimization problem and propose a greedy method by solving the resultant convex optimization problem to select informative channels/kernels. 

\item

\ice{
We propose a new adaptive stopping condition to prevent DCP/DKP from selecting too many channels/kernels when determining the number of selected channels/kernels. \yong{Specifically, the stopping condition incorporates an additional constraint which enforces the number of selected channels to be no more than a predefined value for each layer}.}


\item
Extensive experiments demonstrate the superior performance of our methods on \guo{a variety of} architectures. For example, on ILSVRC-12~\cite{deng2009imagenet}, when pruning 30\% channels from ResNet-50, \jing{DCP improves the original model by 0.36\% \guo{in terms of} Top-1 accuracy.
\guo{To demonstrate the effectiveness of our methods, we also deploy the pruned models on a smartphone (equipped with a Qualcomm Snapdragon 845 processor) \liu{and show significant acceleration on the mobile CPU.}
}
}
\end{itemize}

\jing{
This paper extends our preliminary version~\cite{zhuang2018discrimination} from several aspects.  
1) We propose two training techniques to reduce the computational overhead of DCP while still maintaining comparable or even better performance. 
2) We apply the improved DCP to more compact models (i.e., MobileNetV1 and MobileNetV2) and achieve promising performance on ILSVRC-12. We further deploy the pruned models on a smartphone with a Qualcomm Snapdragon 845 processor to investigate the inference acceleration. 
3) We apply the improved DCP to compress the latest face recognition models.
4) We extend DCP for kernel pruning to further compress models at kernel level. 
5) We propose a new adaptive stopping condition for the optimization. 
6) We provide more ablative studies to investigate the effectiveness of our methods.
}

\section{Related Work}
\noindent\textbf{Network quantization.}
Quantization-based methods represent the network weights and/or activations with very low precision, which yields highly compact DNNs compared to their floating-point counterparts. The extreme case is the binary neural networks (BNNs) where both weights and activations are constrained to $\{+1, -1\}$~\cite{hubara2016binarized, rastegari2016xnor,bulat2018hierarchical}. 
\jing{
In this way, \ice{one} can replace the matrix multiplication operations with the light-weighted bitwise XNOR-popcount operations. As a result, the 1-bit convolutional layer can achieve up to 32$\times$ memory saving and 58$\times$ speedup on CPUs~\cite{rastegari2016xnor, Zhuang_2019_CVPR}.
}
However, BNNs still suffer from significant accuracy decreases.
To \guo{reduce} this accuracy gap, fixed-point methods have been proposed to represent weights and activations with higher bitwidth. Uniform fixed-point approaches \cite{zhou2016dorefa, zhuang2018towards} designed quantizers with a constant quantization step. To improve the precision of the discrete uniform quantizer, \cite{choi2018pact, jung2019learning} explicitly parameterized and optimized the quantization intervals. 

\noindent\textbf{Sparse or low-rank connections.}
To reduce the storage requirements of neural networks, Han~\etal\cite{han2015learning} suggested that neurons with zero input or output connections can be safely removed from the network. With the help of the $\ell_1/\ell_2$ regularization, weights are pushed to zeros during training. Subsequently, the compression rate of AlexNet can reach $35\times$ with the combination of pruning, quantization, and Huffman coding~\cite{han2015deep}. Considering the importance of parameters that are changed during weight pruning, Guo~\etal\cite{guo2016dynamic} proposed dynamic network surgery (DNS). Training with sparsity constraints~\cite{srinivas2017training,wen2016learning} has also been studied to reach a higher compression rate.
Deep models often contain many correlations among channels. To remove such redundancy, low-rank approximation approaches have been widely studied~\cite{denton2014exploiting,gong2014compressing,jaderberg2014speeding,sindhwani2015structured}. For example,  Zhang~\etal\cite{zhang2016accelerating} sped up VGGNet for 4$\times$ with negligible performance degradation on ILSVRC-12. 
However, low-rank approximation approaches are not efficient for \guo{the convolutions with small kernels size, \eg, $1 {\times} 1$ kernel~\cite{TaiXWE15}}.

\begin{figure*}[t]
	\centering
	\includegraphics[width=0.93\textwidth]{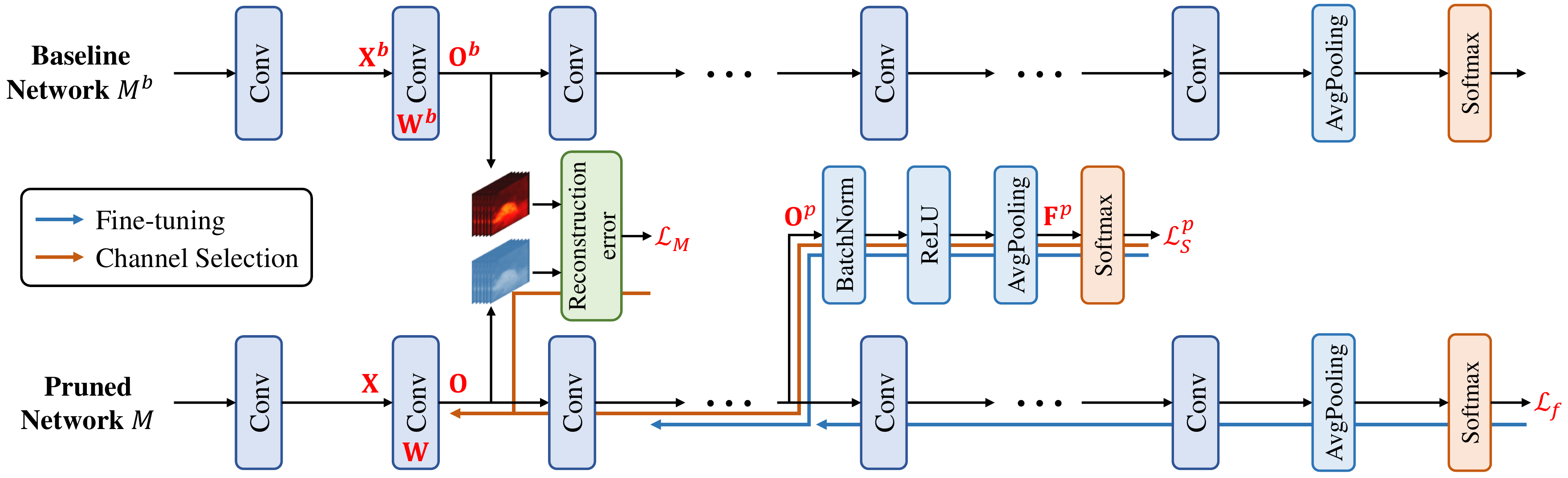}
	\caption{Illustration of discrimination-aware channel pruning. Here, $\mL_S^p$ denotes the discrimination-aware loss (\eg, cross-entropy loss or additive angular margin loss) in the $L_p$-th layer, $\mL_M$ denotes the reconstruction loss, and $\mL_f$ denotes the final loss. \jing{DCP first updates the model $M$ and learns the parameters $\{\btheta^p\}_{p=1}^P$ with $\{\mL^p_S\}_{p=1}^P$ and $\mL_f$. Then, DCP performs channel pruning with $P + 1$ stages. At each stage, for example, in the $p$-th stage, DCP conducts the channel selection for each layer in $\{L_{p-1}+1,\dots,L_p\}$ with the corresponding $\mL_S^p$ and $\mL_M$. }}
	\label{fig:network_architecture}
\end{figure*}

\noindent\textbf{Network pruning.}
Network pruning aims at removing redundant modules, \guo{\eg, channels or kernels,} to \ice{accelerate} the run-time inference. The resultant pruned models would have fewer parameters and lower computational \liu{overhead. }
In order to measure the importance of the network module, different metrics~\cite{li2016pruning,hu2016network,yu2017nisp,he2019filter,lee2018snip,rao2018runtime,zhonghui2019gate,molchanov2019taylor} were proposed.
With a sparsity regularizer in the objective function, training-based methods~\cite{alvarez2016learning,liu2017learning,wen2016learning,zhu2018improving,lin2019towards,Lin2019Toward} were proposed to learn the compact models in the training phase. Considering efficiency, reconstruction-methods~\cite{he2017channel,hu2016network,li2016pruning,luo2018thinet,jiang2018efficient} transformed the channel selection problem into the optimization of the reconstruction error. 
\liu{Recently, several methods~\cite{he2018soft,lin2018accelerating} have been proposed to prune the redundant filters in a dynamic such that the pruned filters can be recovered during training.} 
\guo{Apart from these methods, the pruning rate for each layer can also be automatically determined by reinforcement learning~\cite{he2018amc,tan2019efficientnet}, greedy method~\cite{yang2018netadapt} or evolutionary search~\cite{liu2019metapruning}.} 
\liu{Compared with the proposed methods, most existing methods use heuristic metrics to identify informative modules.} Specifically, both our proposed methods and reconstruction-based methods use the reconstruction error during channel selection.
\jing{However, unlike these methods, \jing{our proposed} DCP/DKP introduce additional losses to select those channels/kernels that actually contribute to the discriminative power of deep networks. \liu{Moreover, compared with those methods that use reinforcement learning\cite{he2018amc} or evolutionary algorithms~\cite{yang2018netadapt},
}  \ice{DCP/DKP use an} adaptive stopping condition to automatically determine the sparsity for each layer.
}

\section{Preliminary}
\label{sec:preliminary}
Let $\{\bx_i, y_i\}_{i=1}^{N}$ be the training samples, where $N$ indicates the number of samples. 
Given an $L$-layer 
deep network $M$, let $\bW \in \mmR^{n\times c\times h_{f}\times z_{f}}$
represents the model parameters \wrt the $l$-th convolutional layer (or block). Here, $h_{f}$ and $z_{f}$ denote the height and width of the filters, respectively; $c$ and $n$ denote the number of \textbf{input} \ice{channels} and \textbf{output} \ice{filters}, respectively. \jing{The parameter $\bW$ contains $n \times c$ kernels \guo{in total}. Each kernel, for example, the \jing{kernel} $\bW_{j,k} \in \mmR^{h_f \times z_f}$ \wrt the $k$-th input channel and $j$-th filter, is a matrix with \guo{the dimension of} $h_f \times z_f$.} 
Let $\bX \in \mmR^{N\times c\times h_{in} \times z_{in}}$ and $\bO\in \mmR^{N\times n \times h_{out} \times z_{out}}$ be the input feature maps and the involved output feature maps, respectively. Here, $h_{in}$ and $z_{in}$ denote the height and width of the input feature maps, respectively; $h_{out}$ and $z_{out}$ represent the height and width of the output feature maps, respectively.
Moreover, let \jing{$\bX_{i,k} \in \mmR^{h_{in} \times z_{in}}$} be the \liu{input} feature map of the $k$-th channel for the $i$-th sample. 
The output feature map \ice{w.r.t. the $j$-th filter of $\bf{W}$ (\ie, ${\bf{W}}_j$)}
for the $i$-th sample, denoted by \jing{$\bO_{i,j} \in \mmR^{h_{out} \times z_{out}}$, is obtained by \liu{performing convolution on the input feature maps $\bX_i$ of the $i$-th sample using the kernel ${\bf{W}}_j$}:
\begin{equation}\label{eq:compute_o}
	\bO_{i,j} = \sum_{k=1}^{c} \bX_{i,k} * \bW_{j,k},
\end{equation}
where $*$ denotes the convolutional operation. }

Given a pre-trained model \liu{$M^b$}, \textbf{Channel Pruning} aims to prune those redundant channels in $\bW$ to reduce the model size and accelerate the inference speed in Eq.~(\ref{eq:compute_o}). 
In order to choose channels, we introduce a variant of the $\ell_{2,0}$-norm\footnote{We can also use other norms to compute the number of selected channels.} $||\bW||_{2,0} = \sum_{k=1}^{c} \Omega(\sum_{j=1}^n ||\bW_{j,k}||_F)$,
where $\Omega(a) = 1$ if $a \neq 0$ and $\Omega(a) = 0$ if $a = 0$, and $||\cdot||_F$ represents the Frobenius norm. In order to induce sparsity, we impose an $\ell_{2,0}$-norm
constraint on $\bW$:
\begin{equation}
\label{eq:l2-zero-norm-channels}
	||\bW||_{2,0} = \sum_{k=1}^{c} \Omega(\sum_{j=1}^n ||\bW_{j,k}||_F) \leq \kappa_{c}^l,
\end{equation}
where $\kappa_{c}^l$ denotes the desired number of channels at layer $l$. Or equivalently, given a predefined pruning rate $\eta \in (0, 1)$, it follows that $\kappa_c^l = \lceil (1-\eta) c\rceil$. \jing{When some channels \guo{of $\bW$} are removed, 
\guo{the computation w.r.t. these channels can be effectively avoided.}
\guo{As a result, the pruned models would have fewer parameters and lower computational costs than the original models.}}

\section{Proposed Method}
\label{sec:proposed_method}

\liu{
Identifying the informative channels, also known as channel selection, is an important problem in channel pruning. 
Most existing methods~\cite{he2017channel,luo2017thinet} conduct channel pruning by minimizing the reconstruction error of feature maps between the pre-trained model and the pruned one. However, merely minimizing the reconstruction error may cause some redundant channels to be mistakenly selected, even though they are actually irrelevant to the discriminative power of the network. 
This issue will be even severer when the network becomes deeper.

In this paper, we highlight that an informative channel, no matter where it is, should contribute to the discriminative power of the network; otherwise, it should be removed. Based on this intuition, we propose a discrimination-aware channel pruning (DCP) scheme to find the channels that actually contribute to the discriminative power. To this end, relying on a pre-trained model, we first introduce multiple discrimination-aware losses into the network to increase the discriminative power of the intermediate layers. Then, we conduct channel selection to select the most discriminative channel by considering both the discrimination-aware loss and the reconstruction error of the feature maps. In the following, we will illustrate the details of our method.
}

\subsection{Motivation}
\label{sec:motivation}
We seek to perform channel pruning by keeping those channels that actually contribute to the discriminative power of the network. In practice, however, it is very difficult to measure the discriminative power of channels due to the complex operations (such as ReLU activation and Batch Normalization) in CNNs. One may consider a channel as an important one if the final loss $\mL_f$ would sharply increase without it.
\jing{
However, for deep models, its shallow layers often have little discriminative power due to the long path of propagation.
As a result, it is not practical to evaluate the discriminative power when the network is very deep. 
}

To increase the discriminative power of the intermediate layers, one can introduce additional losses to the intermediate layers of the deep networks~\cite{szegedy2015going, lee2015deeply, guo2020multi}.
In this paper, we insert $P$ discrimination-aware losses $\{\mL^p_S\}_{p=1}^P$ evenly into the network, as shown in Figure~\ref{fig:network_architecture}. Let $\{L_1, ..., L_P, L_{P+1}\}$ be the layers at which we put the losses, with $L_{P+1}=L$ being the final layer. 
It is worth mentioning that, we can add one loss to each layer of the network, where we have $L_l = l$. However, this can be very computationally expensive yet not necessary.

\subsection{Construction of discrimination-aware loss}
The construction of discrimination-aware loss $\mL^{p}_S$ is very important in our method. As shown in Figure~\ref{fig:network_architecture}, $\mL^{p}_S$ uses the output of layer $L_p$ as the input feature maps. To make the computation of the loss feasible, we impose an average pooling operation over the feature maps. Moreover, to accelerate the convergence, we apply batch normalization~\cite{ioffe2015batch, guo2018double} and ReLU~\cite{nair2010rectified} before performing the average pooling. In this way, the input feature maps for the loss at layer $L_p$, denoted by $\bF^p(\bW)$, can be computed by
\begin{equation}
{\bF}^p(\bW) = {\mathrm{AvgPooling}}(\mathrm{ReLU}(\mathrm{BN}(\bO^p))),
\label{eq:bar_z}
\end{equation}
where $\bO^p$ represents the output feature maps of layer $L_p$. Let $\bF^{(p,i)}$ be the feature maps \wrt the $i$-th example. The discrimination-aware loss \wrt the $p$-th loss is formulated as
\begin{equation}
\mL^p_S(\bW) = - \frac{1}{N} \left[
\sum_{i=1}^N \sum_{t=1}^m I\{y^{(i)}=t\} \log{\frac{e^{{\btheta_t^p}^\top \bF^{(p,i)}}}{\sum_{k=1}^m e^{{\btheta_k^p}^\top \bF^{(p,i)}}}} \right],
\label{eq:compute_ls}
\end{equation}
where $I\{\cdot\}$ is the indicator function, $\btheta^p\in \mmR^{n_p\times m}$ denotes the classifier weights of the fully connected layer, $n_p$ denotes the number of input channels of the fully connected layer and $m$ is the number of classes. \jing{Note that we can also use other losses as the additional loss, such as the additive angular margin loss~\cite{Deng2019}. (See results in Section~\ref{sec:experiment_face_recognition}).}

\subsection{Optimization problem for channel pruning}
Since a pre-trained model contains very rich information about the learning task, similar to~\cite{luo2018thinet}, we hope to reconstruct the feature maps in the pre-trained model
{
by minimizing the reconstruction error of feature maps between the pre-trained model $M^b$ and the pruned one. Formally, the reconstruction error can be measured by the mean squared error (MSE) between the feature maps of the baseline network and the pruned one as follows:
\begin{equation}\label{eq:compute_lmse}
\mL_M(\bW)=\frac{1}{2N\cdot n\cdot h_{out}\cdot z_{out}}\sum_{i=1}^N\sum_{j=1}^n||\bO_{i,j}^b - \bO_{i,j}||_F^2,
\end{equation}
where $\bO_{i,j}^b \in \mmR^{h_{out} \times z_{out}}$ denotes the feature maps of $M^b$.
}
By considering both discrimination-aware loss and reconstruction error, we have a joint loss function as follows:
\begin{equation}
\mL(\bW) = \lambda \mL_M(\bW) + \mL_S^p(\bW),
\label{eq:compute_loss}
\end{equation}
where $\lambda$ balances the two terms. 

\begin{prop}\textbf{\emph{(Convexity of the loss function) }} \label{prop: loss}
	Let $\bW$ be the model parameters of a considered layer. Given the discrimination-aware loss and the mean square loss defined in Eqs.~(\ref{eq:compute_ls}) and~(\ref{eq:compute_lmse}), the joint loss function $ \mL (\bW) $ is convex \wrt $\bW$.\footnote{The proof can be found in Section 7 in~\cite{zhuang2018discrimination}.}
\end{prop}

By introducing the $\ell_{2,0}$-norm constraint, the optimization problem for discrimination-aware channel pruning \guo{becomes}
\begin{equation}
\min_{\bW}~~\mL(\bW),~~~~\st~~ ||\bW||_{2,0}\le \kappa_c^l,
\label{eq:object_func}
\end{equation}
where $\kappa_c^l$ is the number of channels to be selected. In our method, the sparsity of $\bW$ can be either determined by a predefined pruning rate (See Section~\ref{sec:preliminary}) or automatically adjusted by \guo{the adaptive} stopping conditions in Section~\ref{sec:stop_conditions}. 

\begin{algorithm}[t]
	\caption{{\small{Discrimination}-aware channel pruning}}
	\begin{small}
		\begin{algorithmic}[1]
		    \REQUIRE Pre-trained model $M^b$, training data $\{\bx_i, y_i\}_{i=1}^{N}$, and hyperparameters $\{\kappa_c^l\}_{l=1}^L$.
		    \ENSURE Pruned model $M$.
			\STATE Initialize $M$ using $M^b$.
			\STATE Insert losses $\{\mL^p_S\}_{p=1}^P$ to layers $\{L_1, ..., L_P\}$, respectively.
			\STATE Learn $\{\btheta^p\}_{p=1}^P$ and \textbf{Fine-tune} $M$ with $\{\mL^p_S\}_{p=1}^P$ and $\mL_f$.
			\STATE Initialize $M^b$ using $M$.
			\FOR{$p \in \{1, ..., P+1\}$}
			\FOR{$l \in \{L_{p-1}+1,..., L_p\}$}
			\STATE Do \textbf{Channel Selection} for layer $l$ using Algorithm~\ref{alg:channel_selection}.
			\ENDFOR
			\ENDFOR
		\end{algorithmic}
	\end{small}
	\vspace{0.04in}
	\label{alg:dcp}
\end{algorithm}

\subsection{Discrimination-aware channel pruning}
\label{sec:dcp}
By introducing $P$ losses $\{\mL^p_S\}_{p=1}^P$ to the intermediate layers, the proposed discrimination-aware channel pruning (DCP) method is shown in Algorithm~\ref{alg:dcp}. Starting from a pre-trained model $M^b$, \jing{DCP first updates the model $M$ and learns the parameters $\{\btheta^p\}_{p=1}^P$. Then, DCP performs channel pruning with $(P+1)$ stages.} 
Algorithm~\ref{alg:dcp} is called discrimination-aware in the sense that the additional losses and the final loss are considered to fine-tune the model.
Moreover, the additional losses will be used to select channels, as discussed below.

\jing{At the beginning of channel pruning, we first construct additional losses $\{\mL^p_S\}_{p=1}^P$ and insert them at layer $\{L_1, ..., L_P\}$ (See Figure~\ref{fig:network_architecture}).
Next, we learn the parameters $\{\btheta^p\}_{p=1}^P$ and fine-tune the model $M$ at the same time with both the additional losses $\{\mL^p_S\}_{p=1}^P$ and the final loss $\mL_f$. }
During fine-tuning, all the parameters in $M$ are updated. Here, with fine-tuning, the parameters regarding the additional losses can be well learned. 
\footnote{The details of the fine-tuning algorithm can be found in~\cite{guo2020multi}.}
\ice{After doing fine-tuning} with $\{\mL^{p}_S\}_{p=1}^P$ and $\mL_f$, the discriminative power of the intermediate layers can be significantly improved. \jing{Then, we initialize the baseline model $M^b$ with the fine-tuned model $M$ and perform channel pruning with ($P+1$) stages. At each stage, for example, in the $p$-th stage, \ice{we consider the layers of the current stage independently}, and conduct channel selection for the layers in $\{L_{p-1}+1, ..., L_{p}\}$ with corresponding $\mL_S^p$ and $\mL_M$. \liu{
Following~\cite{zhang2016accelerating,he2017channel}, we perform channel selection from the shallower layers to the deeper layers in this paper. 
}}

\begin{algorithm}[t]
	\caption{\small{Greedy algorithm for channel selection}}
	\begin{small}
		\begin{algorithmic}[1]
		    \REQUIRE Training data $\{\bx_i, y_i\}_{i=1}^{N}$, model $M$, hyperparameters $\kappa_c^l$, $\epsilon$.
		    \ENSURE Selected channel index set $\mA$ and model parameters $\bW_{\mA}$.
			\STATE Initialize $\mA_0 \gets \emptyset$, $\bW^0 = {\bf 0}$, and $t=1$.%
			\WHILE{(stopping conditions are not achieved)}
			\STATE Compute gradients of $\mL$ \wrt $\bW^{t-1}$:  $\bG^{t-1}={\partial \mL}/{\partial \bW^{t-1}}$.
			    \STATE Find the $B$ largest $||\bG_{:, k}^{t-1}||_F$ and record their indices in $\mJ_t$. 
				\STATE Let $\mA_{t} \gets \mA_{t-1} \cup \mJ_t$.
				\STATE Solve the following Subproblem to update $\bW_{\mA_{t}}^{t-1}$: 
                \begin{equation}
                    \label{eq:subproblem}
                         \min_{\bW^{t-1}}~~\mL(\bW^{t-1}), ~~\st~~\bW_{\mA_{t}^c}^{t-1} = \0.
                \end{equation}
                \vspace{-0.10in}
			\STATE Set $\bW^{t} \gets \bW^{t-1}$ and let $t\gets t+1$.
			\ENDWHILE
		\end{algorithmic}
		\label{alg:channel_selection}
	\end{small}
\end{algorithm}

\subsection{Greedy algorithm for channel selection}
\label{sec:learning}
Due to the non-convexity of $\ell_{2,0}$-norm, directly optimizing Problem~(\ref{eq:object_func}) is very difficult. To address this issue, following the general greedy methods in~\cite{liu2014forward,bahmani2013greedy,yuan2014gradient,tan2014towards,tan2015matching}, we propose a greedy algorithm to solve Problem~(\ref{eq:object_func}).

\guo{We show the details of the proposed greedy algorithm in Algorithm~\ref{alg:channel_selection}.}
At the beginning of the channel selection, we remove all the channels by setting $\bW^0 = {\bf 0}$. \guo{In} each iteration, we first select those channels that actually contribute to the discriminative power of the network. Then, we solve Subproblem (\ref{eq:subproblem}) with the selected channels only. We will give the details of selecting the channels with discriminative power and the subproblem optimization in the following subsections.

\subsubsection{Discrimination-aware channel selection}
At each iteration of Algorithm~\ref{alg:channel_selection}, we compute the gradients $\bG_{:,k}^{t-1} = \partial \mL / \partial \bW_{:,k}^{t-1}$, where $\bW_{:,k}^{t-1}$ denotes the parameters for the $k$-th input channel at iteration $t$.
Since we set $\bW^0 = {\bf 0}$ at the beginning of channel selection, the initial loss value will be very large. Apparently, selecting any channel at the $t$-th iteration will decrease the loss function, and the channel with the largest gradient $ ||\bG_{:, k}^{t-1}||_F $ will decrease the loss function the most. With the selection criteria, we choose $B$ channels corresponding to the $B$ largest $  ||\bG_{:, k}^{t-1}||_F $ as active channels and record their indices into $\mJ_t$.
Let $\mA_t \subset \{1,\dots, c\}$ be the index set of the selected channels up to iteration $t$, \ie, $\mA_t = \cup_i \mJ_i, i=1,\dots,t$. In general, once a channel is added into $\mJ_t$, it is unlikely to be selected in the following iteration. However, if we do not solve Subproblem (\ref{eq:subproblem}) accurately, some of the selected channels may have a large value of $ ||\bG_{:, k}^{t-1}||_F $, thus they might be chosen again. To avoid this issue, we propose choosing channels from $\{1, \dots, c\} \backslash \mA_{t-1}$ to form $\mJ_{t}$. In this way, there will be no overlapping channels among the $\mJ_i$'s, where $i=1,\dots,t$.

\subsubsection{Subproblem optimization}
Once $\mA_{t}$ is determined, we optimize $\bW^{t-1}$ \wrt the selected channels by minimizing Subproblem (\ref{eq:subproblem}). Here, $\bW_{\mA^c_{t}}^{t-1}$ denotes the subtensor indexed by $\mA^c_{t}$, and $\mA^c_{t}$ is the complementary set of $\mA_{t}$, \ie, $\mA^c_{t} = \{1,\dots, c\} \backslash \mA_{t}$. 
To solve the Subproblem in Eq.~(\ref{eq:subproblem}), we apply stochastic gradient descent (SGD) and update $\bW_{\mA_{t}}^{t-1}$ by
\begin{equation}\label{eq:update_w}
\bW_{\mA_{t}}^{t-1}\gets \bW_{\mA_{t}}^{t-1} - \gamma \frac{\partial\mL}{\partial \bW_{\mA_{t}}^{t-1}},
\end{equation}
where $\bW_{\mA_{t}}^{t-1}$ denotes the subtensor indexed by $\mA_{t}$, and $\gamma$ denotes the learning rate. 

Note that when optimizing Subproblem~(\ref{eq:subproblem}), $\bW_{\mA_{t}}^{t-1}$ is warm-started from the fine-tuned model $M$. As a result, the optimization can be completed very quickly. \jing{Moreover, since \ice{Subproblem} (\ref{eq:subproblem}) is convex with respect to $\bW^{t-1}$ for one layer, we do not need to consider all the data for optimization.} To make a trade-off between the efficiency and performance, we sample a subset of images randomly from the training data for optimization.
\footnote{We explore the number of samples in Section 11 in~\cite{zhuang2018discrimination}.}
Last, since we use SGD to update $\bW_{\mA_t}^{t-1}$, the learning rate $\gamma$ should be carefully adjusted to achieve an accurate solution. 

\subsection{Stopping conditions} \label{sec:sparsity}
\label{sec:stop_conditions}
Given a predefined hyperparameter $\kappa_c^l$ in Problem (\ref{eq:object_func}), Algorithm~\ref{alg:channel_selection} will be stopped if $||\bW^t||_{2,0} > \kappa_c^l$. However, $\kappa_c^l$ is hard to be determined in practice. To solve this issue, we adopt the following two stopping conditions to terminate Algorithm~\ref{alg:channel_selection}.

\noindent \textbf{Stopping condition 1:} 
Since $\mL$ is convex, $\mL(\bW^t)$ will monotonically decrease with the iteration index $t$ in Algorithm~\ref{alg:channel_selection}. Therefore, we can adopt the following stopping condition:
\begin{equation}
\label{eq:stop_con_outer}
\frac{|\mL(\bW^{t-1}) - \mL(\bW^t)|}{{\mL(\bW^0)}} \leq \epsilon,
\end{equation}
where $\epsilon$ is a tolerance value.



\noindent \textbf{Stopping condition 2:} Directly using \textbf{stopping condition 1} can automatically determine the number of selected channels.
However, in practice, if most of the channels are relatively informative and each of them contributes to the loss function, then the objective function value w.r.t. Eq.~(\ref{eq:compute_loss}) may decrease very slowly during the greedy selection process. In this case, \textbf{stopping condition 1} may tend to select more channels or even all channels to achieve sufficiently small loss function value w.r.t. Eq.~(\ref{eq:compute_loss}).
\ice{To address this, we propose \textbf{stopping condition 2} which incorporates an additional constraint to force the number of selected channels to be no more than a predefined value for each layer.}
Specifically, given a minimum pruning rate $\eta_{\text{min}}$, we use the following condition:
\begin{equation}
    \label{eq:stopping_condition_2}
    \begin{split}
        & \frac{|\mL(\bW^{t-1}) - \mL(\bW^t)|}{{\mL(\bW^0)}} \leq \epsilon, ~\text{or}~ \\  
         &  ||\bW^t||_{2,0} > \lceil (1-\eta_{\text{min}}) c\rceil.
    \end{split}
\end{equation}
If the above condition is achieved, the pruning process will be stopped earlier, and the number of selected channels will be automatically determined, \ie, $||\bW^t||_{2,0}$. As a result, the pruned models tend to have lower computational costs.  Note that \textbf{stopping condition 2}  is built on top of \textbf{stopping condition 1}. In practice,  \textbf{stopping condition 1} alone often works well and \textbf{stopping condition 2} is satisfied accordingly.
The comparisons of different stopping conditions are shown in Section~\ref{sec:effect_upper_bound}.
	
\subsection{Techniques for efficient implementations}
	\label{sec:efficient_techniques}
	\guo{For model compression methods, training cost has always been a key factor in real-world applications. Regarding this issue, we propose two methods to improve the training efficiency of DCP.}
    
    \noindent \textbf{Single round fine-tuning}.
    In Algorithm~\ref{alg:dcp}, fine-tuning plays a crucial role in improving the discriminative power of the intermediate layers. However, it leads to high computational costs. In our conference version~\cite{zhuang2018discrimination}, we perform fine-tuning and channel selection stage-wisely. In total, we fine-tune the network $P + 1$ times, which increases the computational cost of channel pruning. To improve the efficiency of channel pruning, we fine-tune the network only once in step 3 of Algorithm~\ref{alg:dcp}, which reduces the times of fine-tuning. Moreover, the fine-tuning process is the same for the network with the same architecture but different pruning rates.
    Therefore, we can store the model after fine-tuning. In this way, the pruned networks with the same architecture but different pruning rates can skip fine-tuning in step 3 of Algorithm~\ref{alg:dcp},
    which greatly reduces the computational cost of channel pruning. We show the effect of single round fine-tuning in Section~\ref{sec:effect_single_shot_finetuning}.
    
	\noindent \textbf{Feature reusing}. 
	In Algorithm~\ref{alg:channel_selection}, we need to compute the input feature maps of layer $l$ to compute the loss function. To obtain the input feature maps, we have to feed $N$ images into the network from layer 1 to layer $l-1$. In \cite{zhuang2018discrimination}, this process is repeated for each iteration, which incurs high computational overhead. Since we do not change the input feature maps during channel selection, we can store and reuse the input feature maps once they have been computed. 
	In this way, we avoid the repeated calculation of the input feature maps, which greatly reduces the computational cost of channel selection. 
	An empirical study on the effect of feature reusing can be found in Section~\ref{sec:effect_feature_reusing}.

\section{Discrimination-aware kernel pruning}
\label{sec:kernel_pruning}
The proposed DCP can select the channels that actually contribute to the discriminative power of the network. 
However, \guo{there exists} a limitation for DCP. Specifically, channel pruning assumes that all kernels in a channel are equally important, which may hamper the performance of the pruned model. In fact, some kernels may not contribute to the discriminative power of the network, resulting in performance degradation of the pruned model. More critically, in DCP, once a channel is pruned (or not selected), even if there may still exist some informative kernels related to it, we are no longer able to find and keep those useful kernels related to the removed channel, which may result in suboptimal performance. 

To solve this issue, we propose a kernel pruning method called discrimination-aware kernel pruning (DKP) to further compress deep networks by removing redundant kernels.
Similar to DCP, we introduce a variant of the $\ell_{2,0}$-norm constraint on $\bW$ to conduct kernel pruning:
\begin{equation}
\label{eq:l2-zero-norm-kernels}
\begin{array}{ll}
||\bW||_{2,0}^{ker} = \sum_{j=1}^n \sum_{k=1}^{c} \Omega(||\bW_{j,k}||_F) \leq \kappa_{ker}^l,
\end{array}
\end{equation}
where $\kappa_{ker}^l$ is the desired number of kernels at layer $l$. When some kernels are removed, the corresponding computational cost w.r.t. the kernels can be effectively reduced.

Starting from a pre-trained model, DKP introduces $P$ additional losses $\{\mL^p_S\}_{p=1}^P$ evenly to the intermediate layers. Then, DKP fine-tunes the model using the addition losses $\{\mL^p_S\}_{p=1}^P$ and the final loss $\mL_f$ to improve the discriminative power of the intermediate layers. \ice{After that}, DKP conducts kernel selection for each layer in a layer/stage-wise manner. 

Similar to DCP, we introduce a variant of the $\ell_{2,0}$-norm constraint into the loss function in Eq. (\ref{eq:compute_loss}). Thus, the optimization problem for DKP can be formulated as:
\begin{equation}
\min_{\bW}~~\mL(\bW),~~~~\st~~ ||\bW||_{2,0}^{ker}\le \kappa_{ker}^l.
\label{eq:optimization_problem_kernel_pruning}
\end{equation}
To solve Problem (\ref{eq:optimization_problem_kernel_pruning}), we propose a greedy algorithm for kernel selection similar to Algorithm \ref{alg:channel_selection}. \liu{At the beginning of the kernel selection, DKP removes all the kernels by setting $\bW^0 = {\bf 0}$.}
Instead of choosing the channels, we choose $B$ kernels with the largest gradients $||\bG_{j, k}^{t-1}||_F = {\partial \mL}/{\partial \bW_{j,k}^{t-1}}$ and put their indices into $\mJ_t$.
Then, we update $\mA_t$ by $\mA_t = \mA_{t-1} \cup \mJ_t$.
Once $\mA_t$ is determined, we optimize $\bW^{t-1}$ \wrt the selected kernels by minimizing Subproblem (\ref{eq:subproblem}), which is similar to DCP. In practice, we may directly apply DKP to compress models (denoted by \textbf{DKP Only}) or sequentially perform kernel selection after performing channel pruning (denoted by \textbf{DCP+DKP}). We investigate the effectiveness of DKP in Section~\ref{sec:effect_dkp}. 

\noindent \textbf{Comparisons with DCP}.
Unlike DCP that focuses on channel selection, DKP seeks to
select the informative kernels related to each channel, making it possible for model compression in a finer way (channel level vs. kernel level). Note that in each greedy optimization step, we can include a relatively large number of kernels (namely, a large $B$). Moreover, similar to DCP, during the subproblem optimization, we do not need to solve the problem exactly. So the complexity of DKP is similar to DCP. In fact, DKP has a comparable training cost with DCP in practice. For example, when pruning 90\% FLOPs from VGGNet on CIFAR-10, the training cost of DKP is 3.25 GPU hours while the cost of DCP is 2.92 GPU hours. More empirical comparisons between DKP and DCP can be found in Section~\ref{sec:effect_dkp}. 


\section{Experiments}
\label{sec:experiments}
\begin{table*}[!t]
	\renewcommand{\arraystretch}{1.3}
	\caption{Performance comparisons on CIFAR-10. "-" denotes that the results are not reported. Top-1 Err. $\uparrow$ is the Top-1 error gap between the pruned model and the baseline model.}
	\label{table:comparision_cifar10}
	\centering
	\scalebox{0.85}{
		\begin{tabular}{c|c||ccc|cc}
			\hline
			Model & Method & \tabincell{c}{Baseline \\Top-1 Err. (\%)} & \tabincell{c}{Pruned \\Top-1 Err. (\%)} & \tabincell{c}{Top-1 \\Err. $\uparrow$ (\%)} & \tabincell{c}{\#Param. $\downarrow$ (\%)} & \tabincell{c}{\#FLOPs $\downarrow$ (\%)} \\
			\hline\hline
			\multirow{11}{*}{\tabincell{c}{ResNet-56}}
			& NISP~\cite{yu2017nisp}      & -    & -    & +0.03 & 42.60 & 43.61 \\
			& ThiNet~\cite{luo2018thinet} & 6.20 & 7.02 & +0.82 & 49.67 & 49.78 \\
			& CP~\cite{he2017channel}     & 7.20 & 8.20 & +1.00 & -    & 50.00 \\
			& AMC~\cite{he2018amc}        & 7.20 & 8.10 & +0.90 & -    & 50.00 \\
			& SFP~\cite{he2018soft}       & 6.41 & 6.65 & +0.24 & -    & 52.60 \\
			& FPGM~\cite{he2019filter}    & 6.41 & 6.51 & +0.10 & -    & 52.60 \\
			& WM$^*$~\cite{howard2017mobilenets}     & 6.26 & 6.76 & +0.50 & 49.67 & 49.78 \\
			& WM+                                & 6.26 & 6.65 & +0.39 & 49.67 & 49.78 \\
            & Random-DCP         & 6.26 & 6.66 & +0.40 & 49.67 & 49.78 \\
            \cdashline{2-7}
            & DCP & 6.26 & 6.28 & +0.02 & 49.67 & 49.78 \\
			& Adapt-DCP               & 6.26 & \textbf{6.23} & \textbf{-0.03} & \textbf{68.48} & \textbf{54.80} \\
			\hline
			\multirow{13}{*}{\tabincell{c}{VGGNet}}
			& PFEC~\cite{li2016pruning}               & 6.75 & 6.60 & -0.15 & 64.00 & 34.18 \\
			& ThiNet~\cite{luo2018thinet}             & 6.01 & 6.15 & +0.14 & 48.29 & 50.08 \\
			& CP~\cite{he2017channel}                 & 6.01 & 6.33 & +0.32 & 48.29 & 50.08 \\
			& Network Slimming~\cite{liu2017learning} & 6.34 & 6.20 & -0.14 & 88.52 & 50.94 \\
			& NRE~\cite{jiang2018efficient}           & 6.54 & 6.60 & +0.06 & 92.70 & 67.60 \\
			& DR~\cite{zhu2018improving}              & 6.42 & 6.69 & +0.27 & 88.30 & 68.63 \\
			& WM$^*$~\cite{howard2017mobilenets}         & 6.02 & 6.39 & +0.37 & 48.29 & 50.08 \\
			& WM+                                         & 6.02 & 6.12 & +0.10 & 48.29 & 50.08 \\
			& Random-DCP            & 6.02 & 5.99 & -0.03 & 48.29 & 50.08 \\
			\cdashline{2-7}
			& DCP       & 6.02 & 5.71 & -0.31 & 48.29 & 50.08 \\
			& Adapt-DCP                       & 6.02 & \textbf{5.45} & \textbf{-0.57} & \textbf{91.69} & \textbf{69.81} \\
			\hline
			\multirow{5}{*}{\tabincell{c}{MobileNetV1}}
			& WM$^*$~\cite{howard2017mobilenets}              & 6.04 & 6.39 & +0.35 & 43.41 & 48.63 \\
			& WM+                                         & 6.04 & 6.35 & +0.31 & 43.41 & 48.63 \\
			& Random-DCP     & 6.04 & 6.36 & +0.32 & 43.41 & 48.63 \\
			\cdashline{2-7}
			& DCP       & 6.04 & 5.54 & -0.50 & 43.41 & 48.63 \\
			& Adapt-DCP                       & 6.04 & \textbf{5.43} & \textbf{-0.61} & \textbf{78.26} & \textbf{66.12} \\
			\hline
			\multirow{5}{*}{\tabincell{c}{MobileNetV2}}
			& WM$^*$~\cite{howard2017mobilenets}        & 5.53 & 5.98 & +0.45 & 23.59 & 27.07 \\
			& WM+                                         & 5.53 & 5.93 & +0.40 & 23.59 & 27.07 \\
			& Random-DCP          & 5.53 & 5.96 & +0.42 & 23.59 & 27.07 \\
			\cdashline{2-7}
			& DCP       & 5.53 & 5.37 & -0.16 & 23.59 & 27.07 \\
			& Adapt-DCP                       & 5.53 & \textbf{5.28} & \textbf{-0.25} & \textbf{40.06} & \textbf{34.44}\\
			\hline
	\end{tabular}}
	\begin{tablenotes}
         \item \footnotesize ~~~~~~~~~~~~~~~~~~~~~~~~~~~$^*$ The results were obtained by our implementations.
    \end{tablenotes}
\end{table*}

To demonstrate the effectiveness of the proposed method, we apply DCP to various architectures, such as ResNet~\cite{he2016deep}, MobileNetV1~\cite{howard2017mobilenets} and MobileNetV2~\cite{sandler2018inverted}, etc.
We conduct experiments on both image classification and face recognition. In order to verify the effectiveness of DKP, we apply DKP to ResNet~\cite{he2016deep} and VGGNet~\cite{simonyanz14a}.
All implementations are based on PyTorch~\cite{paszke2019pytorch}.

We organize the experiments as follows. First, we evaluate the proposed DCP on image classification in Section~\ref{sec:image_classification}. Second, we apply DCP to face recognition in Section~\ref{sec:experiment_face_recognition}. 
Last, we evaluate the proposed DKP in Section~\ref{sec:effect_dkp}.

\begin{table*}[t]
	\renewcommand{\arraystretch}{1.3}
	\caption{Performance comparisons on ILSVRC-12. "-" denotes that the results are not reported. "Top-1 Err. $\uparrow$" and "Top-5 Err. $\uparrow$" denote the Top-1 and Top-5 error gap between the pruned model and the baseline model.}
	\label{table:comparision_ilsvrc12}
	\centering
	\scalebox{0.85}{
		\begin{tabular}{c|c||ccc|ccc|cc}
			\hline
			Model & Method & \tabincell{c}{Baseline \\ Top-1 Err. (\%)} & \tabincell{c}{Pruned \\ Top-1 Err. (\%)} & \tabincell{c}{Top-1 \\ Err. $\uparrow$ (\%)} & \tabincell{c}{Baseline \\ Top-5 Err. (\%)} & \tabincell{c}{Pruned \\ Top-5 Err. (\%)} & \tabincell{c}{Top-5 \\ Err. $\uparrow$ (\%)} & \tabincell{c}{\#Param.  $\downarrow$ (\%)} & \tabincell{c}{\#FLOPs $\downarrow$ (\%)} \\
			\hline\hline
			\multirow{13}{*}{\tabincell{c}{ResNet-50}}
			& SFP~\cite{he2018soft}       & 23.85 & 25.39 & +1.54  & 7.13  & 7.94   & +0.81 & -    & 41.80 \\
			& GAL~\cite{lin2019towards} & 23.85 & 28.05 & +4.20 & 7.13 & 9.06 & +1.93 & 16.86 & 43.03 \\
			& Taylor-FO-BN~\cite{molchanov2019taylor} & 23.82 & 25.50 & +1.68 & - & - & - & 44.53 & 44.99 \\
			& SPP~\cite{wang2018structured} & -   & -     & -      & 8.80  & 9.60   & +0.80 & -    & 50.00 \\
			& CP~\cite{he2017channel}     & -     & -     & -      & 7.80  & 9.20   & +1.40 & -    & 50.00 \\
			& GDP~\cite{lin2018accelerating} & 24.87 & 28.11 & +3.24  & 7.70 & 9.29  & +1.59 & -    & 51.30 \\
			& FPGM~\cite{he2019filter}    & 23.85 &  25.17  & +1.32  & 7.13 & 7.68  & +0.55 & -    & 53.50 \\
			& ThiNet~\cite{luo2018thinet} & 24.70 &  27.97  & +3.27  & 7.80 & 9.01  & +1.21 & 51.55 & 55.50 \\
			& SSR-L2~\cite{Lin2019Toward} & 24.88 &  28.53  & +3.65  & 7.70 & 9.81  & +2.11 & 52.94 & 56.41 \\
			
			& WM$^*$~\cite{howard2017mobilenets} & 23.99 & 26.47 & +2.48 & 7.07 & 8.52 & +1.45 & 51.55 & 55.50 \\
			& WM$^*$(Scratch-B)~\cite{liu2018rethinking} & 23.99 & 25.44 & +1.45 & 7.07 & 8.03 & +0.96 & 51.55 & 55.50 \\
			\cdashline{2-10}
			& DCP       & 23.99 & 25.01 & +1.02 & 7.07 & 7.80  & +0.73 & 51.55 & \textbf{55.50} \\
			& Adapt-DCP       & 23.99 & \textbf{24.85} & \textbf{+0.86} & 7.07  & \textbf{7.70} & \textbf{+0.63} & \textbf{54.99} & 52.41 \\
			\hline
			\multirow{6}{*}{\tabincell{c}{MobileNetV1}}
			& NetAdapt~\cite{yang2018netadapt} & 29.10 & 30.90 & +1.80 & - & -  & - & - & 12.63 \\
			& AMC~\cite{he2018amc} & 29.10 & 29.51 & +0.41 & 10.10 & 10.69 & +0.59 & 43.09 & 48.42 \\
			& MetaPruning~\cite{liu2019metapruning} & 29.10 & 29.60 & +0.50 & - & - & - & - & 50.62 \\
			& WM~\cite{howard2017mobilenets} & 29.10 & 31.60 & +2.50 & 10.10 & 11.80 & +1.70& 38.90 & 42.86 \\
			\cdashline{2-10}
			& DCP & 29.12 & 29.52 & +0.40 & 10.22 & \textbf{10.61} & \textbf{+0.39} & \textbf{45.47} & 49.67 \\
			& Adapt-DCP & 29.12 & \textbf{29.50} & \textbf{+0.38} & 10.22 & 10.78 & +0.56 & 43.27 & \textbf{51.25} \\
			\hline
			\multirow{5}{*}{\tabincell{c}{MobileNetV2}}
			& AMC~\cite{he2018amc} & 28.20 & 29.15 & +0.95 & 9.00 & 10.09  & +1.09 & 33.51 & 30.06 \\
			& MetaPruning~\cite{liu2019metapruning} & 28.20 & 28.80 & +0.60 & - & - & - & - & 27.67 \\ 
		    & WM~\cite{sandler2018inverted} & 28.20 & 30.20 & +2.00 & 9.00 & 10.40 & +1.40& 25.02& 40.52\\
			\cdashline{2-10}
			& DCP & 28.12 & 28.79 & +0.67 & 9.70 & 10.01 & +0.31 & 34.24 & \textbf{32.21} \\
			& Adapt-DCP & 28.12 & \textbf{28.61} & \textbf{+0.49} & 9.70 & \textbf{9.91} & \textbf{+0.21} & \textbf{35.01} & 30.67 \\
			\hline
	\end{tabular}}
	\begin{tablenotes}
         \item \footnotesize $^*$ The results were obtained by our implementations.
    \end{tablenotes}
\end{table*}

\subsection{Experiments on image classification}
\label{sec:image_classification}

\subsubsection{Compared methods}
To investigate the effectiveness of the proposed methods, we include the following methods for study:
\textbf{DCP:} The proposed channel pruning method with a pre-defined pruning rate $\eta$.
\textbf{Adapt-DCP:} DCP with adaptive \jing{\textbf{stopping condition 2}} introduced in Section~\ref{sec:stop_conditions}.
\textbf{WM:} We shrink the width of a network by a fixed ratio and train it from scratch, which is known as the width multiplier~\cite{howard2017mobilenets}.
\textbf{WM+:} Based on WM, we evenly insert additional losses into the network and train it from scratch.
\textbf{Random-DCP:} Relying on DCP, we randomly choose channels instead of using the gradient-based strategy introduced in Algorithm~\ref{alg:channel_selection}. 

We also consider several state-of-the-art channel pruning methods for comparison. On CIFAR-10, we compare DCP with NISP~\cite{yu2017nisp}, ThiNet~\cite{luo2018thinet}, CP~\cite{he2017channel}, AMC~\cite{he2018amc}, SFP~\cite{he2018soft}, FPGM~\cite{he2019filter}, PREC~\cite{li2016pruning}, Network Slimming~\cite{liu2017learning}, NRE~\cite{jiang2018efficient}, and DR~\cite{zhu2018improving}. On ILSVRC-12, we compare DCP with SFP~\cite{he2018soft}, GAL~\cite{lin2019towards}, Taylor-FO-BN~\cite{molchanov2019taylor}, SPP~\cite{wang2018structured}, CP~\cite{he2017channel}, GDP~\cite{lin2018accelerating}, FPGM~\cite{he2019filter}, ThiNet~\cite{luo2018thinet}, SSR-L2~\cite{Lin2019Toward}, NetAdapt~\cite{yang2018netadapt}, AMC~\cite{he2018amc}, and MetaPruning~\cite{liu2019metapruning}. \liu{Following~\citep{liu2017learning,luo2018thinet}, we measure the computational cost of the pruned models by the \yong{number of} floating point operations (FLOPs).}

\subsubsection{Datasets and implementation details}
\label{sec:dcp_details}
We evaluate the proposed DCP on two image classification datasets, including CIFAR-10~\cite{krizhevsky2009learning} and ILSVRC-12~\cite{deng2009imagenet}. CIFAR-10 consists of 50k training samples and 10k testing images with 10 classes. ILSVRC-12 contains 1.28 million training samples and 50k testing images for 1,000 classes. 

Based on the pre-trained model, we apply our method to select informative channels. We first introduce additional losses to increase the discriminative power of the intermediate layers. In practice, we determine the number of additional losses according to the depth of the network
(See Section~\ref{sec:explore_num_losses}).
Specifically, we insert 3 additional losses to ResNet-50 and ResNet-56, and 2 additional losses to VGGNet, ResNet-18, MobileNetV1, and MobileNetV2. 
Then, we fine-tune the model with both the additional losses and the final loss. 
On CIFAR-10, we fine-tune 100 epochs using a mini-batch size of 128. The learning rate starts from 0.1 and is divided by 10 at epochs 40 and 60. On ILSVRC-12, we fine-tune 60 epochs using a mini-batch size of 256. The learning rate is initialized to 0.01 and divided by 10 at epochs 36, 48, and 54.

\ice{After doing fine-tuning}, we conduct channel selection in a stage-wise manner by considering the corresponding additional loss and the reconstruction error of the feature maps.
For ResNet-56 and ResNet-50, we set $\eta$ and $\eta_{min}$ to 0.5 and 0.4, respectively. For VGGNet, MobileNetV1 and MobileNetV2, $\eta$ and $\eta_{min}$ are set to 0.3 and 0.2, respectively. In our experiment, $\lambda$ and $B$ are set to 1.0 and 2.0, respectively.

\ice{After doing channel selection}, we fine-tune the whole network with the selected channels only. 
We use SGD with nesterov~\cite{nesterov1983method} for optimization. The momentum and weight decay are set to 0.9 and $1 \times 10^{-4}$, respectively. On CIFAR-10, we fine-tune 400 epochs using a mini-batch size of 128. The learning rate is initialized to 0.1 and divided by 10 at epochs 160 and 240. 
For ResNet-18 and ResNet-50 on ILSVRC-12, we fine-tune the network for 60 epochs with a mini-batch size of 256. The learning rate starts at 0.01 and is divided by 10 at epochs 36, 48 and 54.
For MobileNetV1 and MobileNetV2 on ILSVRC-12, we fine-tune for 150 epochs and 250 epochs with a mini-batch size of 256. Following~\cite{howard2017mobilenets, sandler2018inverted}, we set the weight decay to $4 \times 10^{-5}$. The learning rate is initialized to 0.09 and decreased to 0 following the cosine function~\cite{loshchilov2016sgdr}.

\subsubsection{Comparisons on CIFAR-10}
We apply DCP to prune ResNet-56, VGGNet, MobileNetV1, and MobileNetV2 and \guo{compare} the performance on CIFAR-10. We report the results in Table~\ref{table:comparision_cifar10}.

From Table~\ref{table:comparision_cifar10}, we have the following observations. First, the models pruned by DCP significantly outperform those pruned by Random-DCP. For example, on VGGNet, DCP reduces the error by 0.28\% compared with Random-DCP, which implies the effectiveness of the proposed channel selection strategy. Second, the inserted additional losses improve the performance of the networks. Specifically, WM+ of VGGNet achieves better performance than WM. Third, our proposed DCP shows much better performance than WM+. For example, on VGGNet, DCP outperforms WM+ by 0.41\% on the testing accuracy. Fourth, compared with DCP, Adapt-DCP further improves the performance of the pruned models with much fewer parameters and FLOPs. Specifically, when applying Adapt-DCP on VGGNet, the pruned model with 91.69\% and 69.81\% reductions in parameters and FLOPs even lowers the testing error by 0.26\% compared with DCP. 

Compared with several state-of-the-art methods, our method achieves the best performance. For example, ThiNet~\cite{luo2018thinet} prunes VGGNet by 48.29\% of the parameters and 50.08\% of the FLOPs with a 0.14\% drop in terms of the Top-1 accuracy. In contrast, our proposed DCP achieves the same speedup ratio with a \textbf{0.31\%} improvement in terms of the Top-1 accuracy. Moreover, for VGGNet, Adapt-DCP outperforms NRE~\cite{jiang2018efficient} and DR~\cite{zhu2018improving} and obtains a \textbf{91.69\%} reduction in the model size and \textbf{69.81\%} in FLOPs. These results show the superior performance of our proposed DCP and Adapt-DCP.

\begin{table}[!t]
	\renewcommand{\arraystretch}{1.3}
	\caption{Inference acceleration of the pruned models on ILSVRC-12. We report the forward time tested on a mobile CPU (Qualcomm Snapdragon 845).}
	\label{table:acceleration_on_arm}
	\centering
	\scalebox{0.78}{
		\begin{tabular}{c||ccccc}
			\hline
			Network & $\eta$ & \#Param. (M) & \#FLOPs (M) & \tabincell{c}{Mobile CPU \\Time (ms)} & \tabincell{c}{Speedup \\ Ratio} \\ 
			\hline\hline
			\multirow{2}{*}{\tabincell{c}{MobileNetV1}} & 0.00 & 4.23 & 1132.44 & 107.25 & 1.00 \\
			 & 0.30 & \textbf{2.31} & \textbf{569.92} & \textbf{55.65} & \textbf{1.93} \\
			\hline
			\multirow{2}{*}{\tabincell{c}{MobileNetV2}} & 0.00 & 3.50 & 594.87 & 62.88 & 1.00 \\
			 & 0.30 & \textbf{2.30} & \textbf{403.24} & \textbf{44.25} & \textbf{1.42} \\
			\hline
		\end{tabular}
	}
\end{table}

\begin{table*}[!t]
	\renewcommand{\arraystretch}{1.3}
	\caption{Performance comparisons of different methods on face recognition. ``VR'' refers to Verification TAR (True Accepted Rate) and ``FAR$10^{-6}$'' refers to the False Accepted Rate at $10^{-6}$. We do not report the inference time of SphereFace and CosFace due to its large model size and computational costs. }
	\label{table:prune_mobilefacenet_lfw}
	\centering
	\scalebox{0.90}{
		\begin{tabular}{c||ccc|c|cc|cccc}
			
			\hline
			\multirow{2}[0]{*}{Model} & \multicolumn{3}{c|}{Validation Accuracy (\%)} & \multirow{1}{*}{VR@FAR$10^{-6}$ (\%)} & \multirow{2}[0]{*}{\#Param. (M)} & \multirow{2}[0]{*}{\#FLOPs (G)} & \multirow{2}[0]{*}{\tabincell{c}{Mobile CPU\\ Times (ms)}} & \multirow{2}[0]{*}{\tabincell{c}{Speedup\\Ratio}} \\
			\cline{2-5}
			& LFW & CFP-FP & AgeDB-30 & MegaFace & & \\
			\hline\hline
			SphereFace~\cite{Schroff_2015_CVPR} & 99.76 & 93.70 & 97.56 & 97.43 & 65.16 & 24.17 & - & - \\
			CosFace~\cite{wang2018cosface} & 99.80 & 94.40 & 97.91 & 98.36 & 65.16 & 24.17 & - & - \\
			\cdashline{1-11}
			LResNet34E-IR~\cite{Deng2019} & 99.72 & 96.39 & 98.03 & 96.71 & 31.81 & 7.10 & 391.52 & 1.00 \\
			LResNet34E-IR (prune 25\% channels) & 99.70 & \textbf{96.71} & 97.63 & \textbf{97.03} & 27.12 & 5.35 & 308.72 & 1.27 \\
			LResNet34E-IR (prune 50\% channels) & 99.70 & 95.90 & 97.70 & 96.14 & \textbf{22.42} & \textbf{3.60} & \textbf{232.61} & \textbf{1.68} \\
			\cdashline{1-11}
			MobileFaceNet~\cite{chen2018mobilefacenets} & 99.50 & 92.23 & 95.63 & 90.38 & 1.00& 0.44 & 36.90 & 1.00 \\
			MobileFaceNet (prune 25\% channels) & 99.38 & 92.17 & 95.47 & 90.33 & \textbf{0.79} & \textbf{0.33} & \textbf{28.71} & \textbf{1.29} \\
			\hline
		\end{tabular}
	}
\end{table*}

\subsubsection{Comparisons on ILSVRC-12}
To verify the effectiveness of the proposed method on the large-scale dataset, we evaluate our method on ILSVRC-12 and report the Top-1 and Top-5 errors with single view evaluation in Table~\ref{table:comparision_ilsvrc12}.
We first apply DCP to compress ResNet-50.
\jing{
Compared with WM (Scratch-B) \cite{liu2018rethinking}, which leads to 1.45\% increase in terms of the Top-1 error, DCP only results in \textbf{1.02\%} degradation in terms of the Top-1 accuracy, which demonstrates the effectiveness of \liu{the pruning scheme in} DCP.
}
Compared with FPGM~\cite{he2019filter}, DCP achieves 0.30\% improvement in terms of the Top-1 accuracy with a 55.50\% reduction in FLOPs. More critically, the model pruned by Adapt-DCP with the smaller model size even outperforms DCP by 0.16\% and 0.10\% on Top-1 and Top-5 accuracy, respectively.

We also apply DCP to prune compact and efficient neural networks, such as MobileNetV1~\cite{howard2017mobilenets} and MobileNetV2~\cite{sandler2018inverted}.
Compared with AMC~\cite{he2018amc} and MetaPruning~\cite{liu2019metapruning}, our proposed DCP achieves better performance. For example, on MobileNetV2, DCP outperforms AMC~\cite{he2018amc} by 0.36\% in terms of the Top-1 accuracy. Moreover, Adapt-DCP pruned MobileNetV2 outperforms MetaPruning~\cite{liu2019metapruning} by 0.11\% in terms of the Top-1 accuracy. These results demonstrate the effectiveness of DCP and Adapt-DCP.

\subsubsection{Model deployment to mobile devices}
\label{sec:realistic_acceleration}
We further deploy the pruned models to a mobile device.
We perform the evaluations on a Xiaomi 8 smartphone, which is equipped with a 2.8 GHz Qualcomm Snapdragon 845 mobile processor. The test-phase computation is carried out on a single large CPU core without GPU acceleration. We report the results in Table~\ref{table:acceleration_on_arm}. 

Compared with the pre-trained model, MobileNetV1 with a pruning rate of 30\% achieves nearly $2 \times$ acceleration on the mobile phone. Moreover, the execution of our pruned MobileNetV2 only requires 44.25ms, \liu{which is much lower than the pre-trained one. These results show that our methods are able to compress the compact models to significantly reduce the inference time. }


\subsection{Experiments on Face Recognition}
\label{sec:experiment_face_recognition}

\subsubsection{Compared methods}
We apply the proposed DCP to LResNet34E-IR~\cite{Deng2019} and MobileFaceNet~\cite{chen2018mobilefacenets} on face recognition. To evaluate the proposed DCP method, we consider several face recognition models for comparison, including SphereFace~\cite{liu2017sphereface} and CosFace~\cite{wang2018cosface}.

\begin{table}[!t]
	\renewcommand{\arraystretch}{1.3}
	\caption{\liu{Comparisons of DCP and DKP} on CIFAR-10. The Top-1 error (\%) of the pre-trained ResNet-56 and VGGNet are 6.26 and 6.02, respectively.  }
	\label{table:comparision_dcp_dkp_cifar}
	\centering

    \scalebox{0.80}{
		\begin{tabular}{c|c||cc|cc}
			\hline
			Model & Method & \tabincell{c}{Pruned \\Top-1 Err. (\%)} & \tabincell{c}{Top-1 \\Err. $\uparrow$ (\%)} & \tabincell{c}{\#Param. \\ $\downarrow$ (\%)} & \tabincell{c}{\#FLOPs \\ $\downarrow$ (\%)} \\
			\hline\hline
			\multirow{8}{*}{\tabincell{c}{ResNet-56}}
			& DCP & 6.28 & +0.02 & 49.67 & 49.78 \\
			& DCP+DKP & 6.23 & -0.03 & \textbf{51.53} & \textbf{50.59} \\
			& DKP Only & \ice{\textbf{6.18}} & \ice{\textbf{-0.08}} & 49.52 & 50.00 \\
			\cdashline{2-6}
			& DCP & 7.02 & +0.76 & 68.29 & 68.44 \\
			& DCP+DKP & 6.96 & +0.70 & \textbf{68.62} & \textbf{68.84} \\
			& DKP Only & \ice{\textbf{6.71}} & \ice{\textbf{+0.45}} & 67.78 & 68.20 \\
			\cdashline{2-6}
			& Adapt-DCP & 6.23 & -0.03 & 68.48 & 54.80 \\
			& Adapt-DKP & \textbf{6.17} & \textbf{-0.08} & \textbf{68.72} & \textbf{58.42} \\
			\hline
			\multirow{8}{*}{\tabincell{c}{VGGNet}}
			& DCP & 6.24 & +0.22 & 72.01 & 74.30 \\
			& DCP+DKP & 6.17 & +0.15 & 73.16 & 74.26 \\
			& DKP Only & \ice{\textbf{6.15}} & \ice{\textbf{+0.13}} & \textbf{74.33} & \textbf{74.54} \\
			\cdashline{2-6}
			& DCP & 7.98 & +1.96 & 88.42 & 90.24 \\
			& DCP+DKP & 7.74 & +1.72 & 89.39 & 90.42 \\
			& DKP Only & \ice{\textbf{7.52}} & \ice{\textbf{+1.50}} & \textbf{90.41} & \textbf{90.50} \\
			\cdashline{2-6}
			& Adapt-DCP & \textbf{5.45} & \textbf{-0.57} & 91.69 & 69.81 \\
			& Adapt-DKP & 5.48 & -0.54 & \textbf{91.90} & \textbf{76.48} \\
			\hline
	\end{tabular}}
\end{table}

\subsubsection{Datasets and implementation details}
We evaluate the proposed DCP method on four benchmark datasets, including LFW~\cite{huang2007labeled}, CFP-FP~\cite{sengupta2016frontal}, AgeDB-30~\cite{moschoglou2017agedb}, and MegaFace~\cite{kemelmacher2016megaface}. 
LFW~\cite{huang2007labeled} contains 13,233 face images from 5,749 identities.
CFP~\cite{sengupta2016frontal} consists of 500 identities, each with 10 frontal and 4 profile images. 
AgeDB-30~\cite{moschoglou2017agedb} contains 12,240 images of 440 identities.
MegaFace~\cite{kemelmacher2016megaface} is a very challenging benchmark dataset to evaluate the performance of face recognition methods at a scale of million distractors.

We use the refined MS-Celeb-1M~\cite{guo2016ms} released by~\cite{Deng2019} for training. The training dataset consists of 5.8M face images from 85k individuals. With the same settings as~\cite{Deng2019}, we first train LResNet34E-IR~\cite{Deng2019} and MobileFaceNet~\cite{chen2018mobilefacenets} from scratch. 
Then, we apply our proposed DCP to compress the pre-trained models.

Before channel pruning, we first insert 2 additive angular margin losses~\cite{Deng2019} into LResNet34E-IR and MobileFaceNet. Then, we fine-tune 15 epochs with both the discrimination-aware losses and the final loss. 
The learning rate is initialized to 0.01 and divided by 10 at epochs 4, 8, and 12. 

\ice{After doing fine-tuning}, we perform channel selection to select the informative channels. \ice{After doing channel selection}, we fine-tune the whole network for 28 epochs. The learning rate is initialized to 0.01 and divided by 10 at epochs 8, 16 and 24. We use SGD with a mini-batch size of 512 for optimization. 

\subsubsection{Performance comparisons}
We report the results in Table~\ref{table:prune_mobilefacenet_lfw}. From the results, we observe that the pruned models with small pruning rates achieve nearly the same performance as the pre-trained model. For example, for LResNet34E-IR, the pruned model with a pruning rate of 25\% even outperforms the pre-trained model on CFP-FP and MegaFace. Moreover, for MobileFaceNet, the pruned model achieves comparable performance as the pre-trained model with only 0.79M parameters and 28.71ms for inference, which is suitable for resource-limited devices. 

Compared with SphereFace~\cite{Schroff_2015_CVPR} and CosFace~\cite{wang2018cosface}, our pruned LResNet34E-IR achieves comparable performance with a much smaller number of parameters and FLOPs. Even with a pruning rate of 50\%, our pruned LResNet34E-IR still achieves comparable performance to the pre-trained model. These results demonstrate the effectiveness of the proposed DCP on face recognition.

\begin{table}[!t]
	\renewcommand{\arraystretch}{1.3}
	\caption{\liu{Comparisons of DCP and DKP} on ILSVRC-12. The Top-1 and Top-5 error (\%) of the pre-trained ResNet-18 are 30.36 and 11.02, respectively. 
	}
	\label{table:comparision_dcp_dkp_cifar_ilsvrc12}
	\centering

	\scalebox{0.85}{
		\begin{tabular}{c|c||cc|cc}
			\hline
			Model & Method & \tabincell{c}{Top-1 \\ Err. (\%)} & \tabincell{c}{Top-5 \\ Err. (\%)} & \tabincell{c}{\#Param. \\ $\downarrow$ (\%)} & \tabincell{c}{\#FLOPs \\ $\downarrow$ (\%)} \\
			\hline\hline
			\multirow{3}{*}{\tabincell{c}{ResNet-18}}
			& DCP & 32.64  & 12.38  & 47.01 & 46.22\\
			& DCP+DKP & 32.57 & 12.20 & \textbf{47.31} & 46.40 \\
			& DKP Only & \textbf{32.46}  & \textbf{12.04}  & 47.12 & \textbf{46.56} \\
			\hline
	\end{tabular}}
\end{table}

\begin{figure}[!t]
	\centering
    \subfigure[Accuracy reduction \liu{vs.} parameter reduction]
    { 
		\label{fig:acc_params} 
		\includegraphics[height=1.6in]{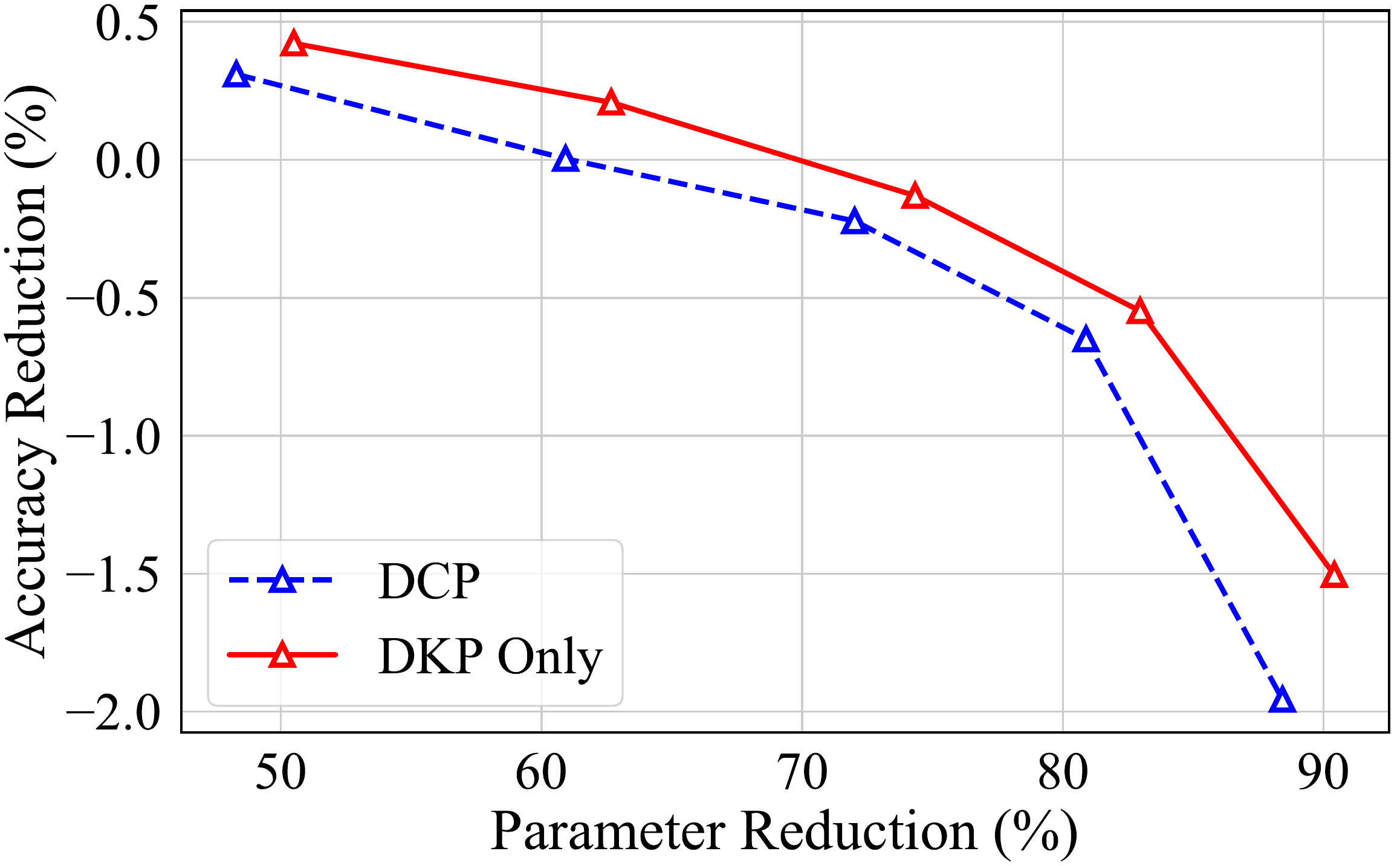}
	} 
	\subfigure[Accuracy reduction \liu{vs.}. FLOPs reduction]
    { 
		\label{fig:acc_flops} 
		\includegraphics[height=1.6in]{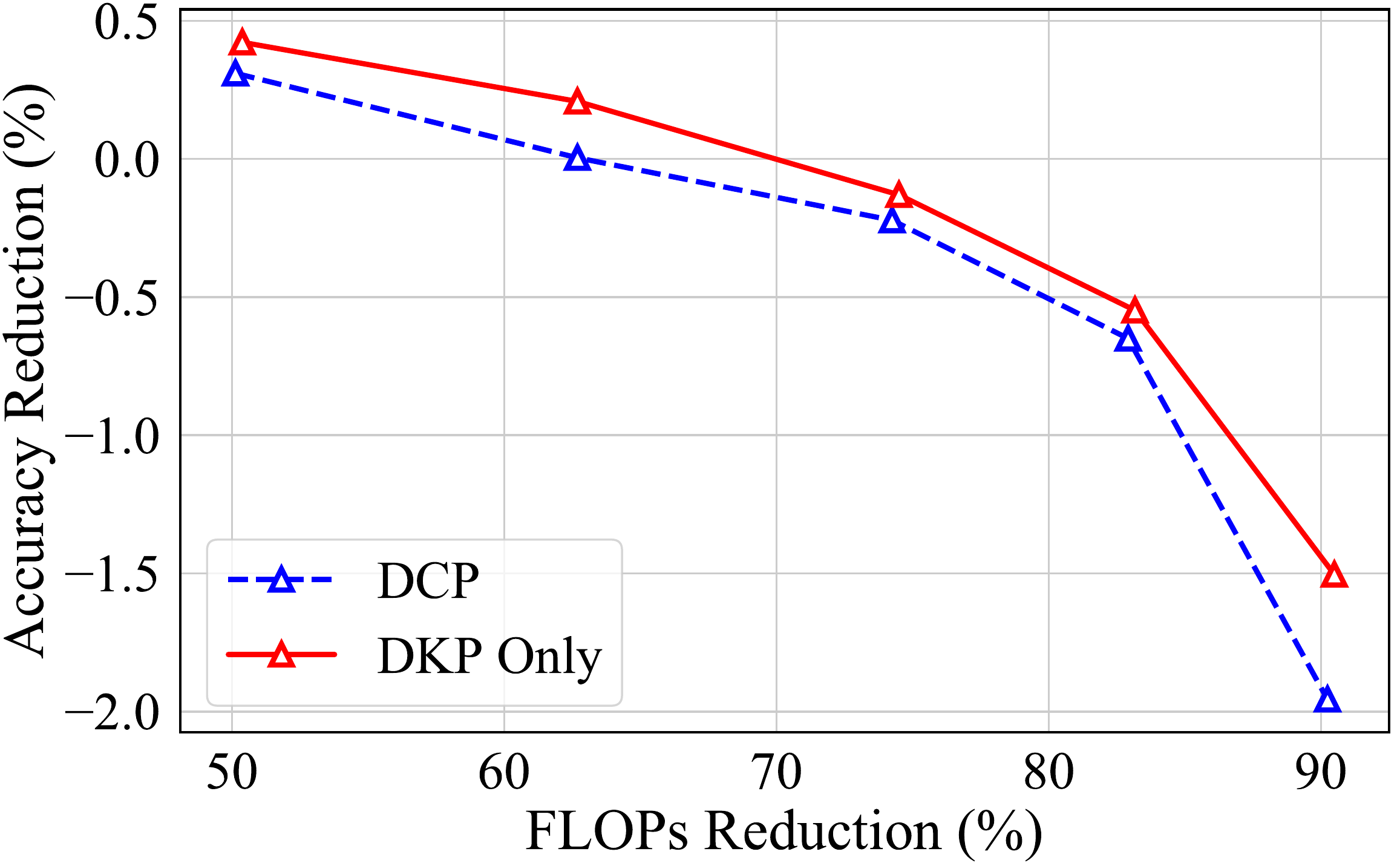}
	}
    \caption{Comparisons of  DCP and DKP in terms of Top-1 accuracy reduction \liu{vs.} parameter/FLOPs reduction. We apply DCP/DKP to prune VGGNet  on CIFAR-10 with different pruning rates. 
    }
    \label{fig:acc_vs_flops_params}
\end{figure}
\subsection{Effectiveness of DKP}
\label{sec:effect_dkp}
\subsubsection{Compared methods}
To investigate the effectiveness of DKP, apart from DCP and Adapt-DCP, we include the following methods for comparisons. 
\textbf{DKP Only}: The proposed kernel pruning method with a fixed pruning rate.
\ice{\textbf{DCP+DKP}: We sequentially perform channel selection and kernel selection.} 
 \jing{\textbf{Adapt-DKP:} DKP Only with adaptive \textbf{stopping condition 2.}}

\subsubsection{Datasets and implementation details}
We evaluate the performance of the pruned models on CIFAR-10 and ILSVRC-12.
For kernel selection, we use the same additional losses introduced in DCP. We set $B$ to a value that is equal to the number of kernels in a channel.
\ice{After doing kernel selection}, we fine-tune the whole network with the selected kernels only.

\subsubsection{Performance comparisons}
We apply DKP to select informative kernels from ResNet-56, VGGNet, and ResNet-18. 
\ice{From Table~\ref{table:comparision_dcp_dkp_cifar} and Table~\ref{table:comparision_dcp_dkp_cifar_ilsvrc12}, directly adopting DKP to prune models achieve the best performance on both CIFAR-10 and ILSVRC-12. For example, for ResNet-18, DKP outperforms DCP by 0.18\% in terms of the Top-1 accuracy with 47.12\% and 46.56\% reductions in the parameters and FLOPs. Moreover, with Adapt-DKP, the resultant models obtain comparable performance but lower computational costs than the ones of Adapt-DCP. Note that the improvement of DCP+DKP over DCP is smaller than that of DKP Only. In DCP+DKP, we simply perform kernel selection after doing channel selection. As a result, the performance highly depends on DCP. Specifically, once a channel is pruned (or not selected), even if there may still exist some informative kernels related to it, we are no longer able to find and keep those useful kernels related to the removed channel, which may result in suboptimal performance and the marginal improvement over DCP.
}
From Figure~\ref{fig:acc_vs_flops_params}, DKP Only consistently outperforms DCP under different pruning rates. 
To be specific, 
VGGNet pruned by DKP Only with a 90.41\% reduction in parameters reduces the Top-1 error by 0.46\% compared with the one pruned by DCP. 
These results demonstrate the effectiveness of DKP. 

\section{Ablation studies}
\jing{
In this section, we conduct ablative studies for the proposed DCP. 
1) We investigate the effect of using different pruning rates in Section~\ref{sec:effect_of_pruning_rate}. 2) We explore the effect of using different $\lambda$ in Section~\ref{sec:effect_of_lambda}. 3) We explore the effect of efficient training strategies in Section~\ref{sec:effective_training}. 
4) We explore the effect of using different $B$ in Section~\ref{sec:effect_B}.
5) We discuss the effect of the tolerance $\epsilon$ in the proposed stopping conditions in Section~\ref{sec:effect_eps}. 
6) We compare different stopping conditions in Section~\ref{sec:effect_upper_bound}.
7) We visualize the feature maps \wrt the pruned/selected channels of ResNet-18 in Section~\ref{sec:visualize_feature}.
8) We explore the influence of the number of additional losses in Section~\ref{sec:explore_num_losses}. 
} 

\begin{table}[!t]
	\renewcommand{\arraystretch}{1.3}
	\caption{Comparisons on ResNet-18 and ResNet-50 with different pruning rates. We report the Top-1 and Top-5 error (\%) on ILSVRC-12.}
	\label{table:results_imagenet_resnet18_resnet50}
	\centering
	\scalebox{0.88}{
		\begin{tabular}{c||c|cc|cc}
			\hline
			Model & $\eta$ & \tabincell{c}{Top-1 \\ Err. (\%)} & \tabincell{c}{Top-5 \\ Err. (\%)} & \tabincell{c}{\#Param. \\ $\downarrow$ (\%)} & \tabincell{c}{\#FLOPs \\ $\downarrow$ (\%)} \\ 
			\hline\hline
			\multirow{4}{*}{ResNet-18} & 0\% & 30.36 & 11.02 & - & - \\
			& 30\% & \textbf{30.77} & \textbf{11.03} & 28.05 & 27.49 \\
			& 50\% & 32.64 & 12.38 & 47.01 & 46.22 \\
			& 70\% & 35.89 & 14.22 & \textbf{65.70} & \textbf{64.25} \\
			\hline
			\multirow{4}{*}{ResNet-50} & 0\% & 23.99&7.07 & - & - \\
			& 30\% & \textbf{23.63} & \textbf{6.96} & 33.40 & 35.72 \\
			& 50\% & 25.01 & 7.80 & 51.55 & 55.50 \\
			& 70\% & 27.28 & 8.83 & \textbf{65.90} & \textbf{71.09} \\ 
			\hline
		\end{tabular}
	}
\end{table}

\begin{table}[!t]
	\renewcommand{\arraystretch}{1.3}
	\caption{Pruning results of ResNet-56 and MobileNetV1 with different $\lambda$. We report the testing error (\%) on CIFAR-10. }
	\label{table:explore_lambda_resnet56}
	\centering
	\scalebox{0.88}{
	\begin{tabular}{c||ccccccc}
		\hline
		$\lambda$ & 0 & 0.1 & 0.5 & 1 & 1($\mL_M$ only) & 5 & 10 \\
		\hline\hline
		ResNet-56 & 5.94 & 5.96 & 5.87 & \textbf{5.80} & 5.99 & 5.85 & 5.93 \\
		MobileNetV1 & 5.66 & 5.46 & 5.59 & 5.54 & 5.43 & \textbf{5.42} & 5.56 \\
		\hline
	\end{tabular}
	}
\end{table}

\subsection{Performance with different pruning rates}
\label{sec:effect_of_pruning_rate}
To study the effect of using different pruning rates $\eta$, we prune 30\%, 50\%, and 70\% channels from ResNet-18 and ResNet-50, and evaluate the pruned models on ILSVRC-12. The experimental results are shown in Table~\ref{table:results_imagenet_resnet18_resnet50}. 

From the results, the pruned models perform worse with the increase of the pruning rate. 
\jing{
However, our pruned ResNet-50 with a pruning rate of 30\% outperforms the pre-trained model with \textbf{0.36\%} and \textbf{0.11\%} improvement in terms of the Top-1 and Top-5 accuracy, respectively.} \liu{It may attribute that DCP serves as a regularization as it effectively reduces model redundancy by selecting the most discriminative channels for each layer.}
Additionally, the performance degradation of ResNet-50 is smaller than that of ResNet-18 under the same pruning rate. 
\jing{
For example, when pruning 50\% of the channels, it only leads to 1.02\% increase in terms of the Top-1 error for ResNet-50. In contrast, it results in 2.28\% increase of the Top-1 error for ResNet-18.
}
It can be attributed to that, compared with ResNet-18, ResNet-50 is more redundant with more parameters, thus it is easier to be pruned.

\subsection{Effect of the trade-off hyperparameter $\lambda$}
\label{sec:effect_of_lambda}
We prune 30\% channels of ResNet-56 and MobileNetV1 on CIFAR-10 with different $\lambda$ values. We report the testing error in Table~\ref{table:explore_lambda_resnet56}. 
\jing{
From the table, the performance of the pruned models first improves and then degrades with the increasing $\lambda$. 
}
Here, a larger $\lambda$ implies that more emphasis is placed on the reconstruction error (See Eq. (\ref{eq:compute_loss})). This demonstrates the effectiveness of the discrimination-aware strategy for channel selection. It is worth mentioning that both the reconstruction error and the cross-entropy loss contribute to better performance of the pruned models, which strongly supports the motivation to select the important channels by $\mL_S^p$ and $\mL_M$. \jing{Note that setting $\lambda$ to 1.0 does not lead to the best performance considering different architectures and datasets. For simplicity, we set $\lambda$ to 1.0 by default in our experiments.}

\begin{table}[!t]
	\renewcommand{\arraystretch}{1.3}
    \caption{Effect of efficient single round fine-tuning. We report the testing error (\%) and the time of channel pruning on CIFAR-10 and ILSVRC-12.}
	\label{table:comparison_NIPS_TPAMI}
	\centering
	\scalebox{0.88}{
		\begin{tabular}{c|c||cc|c}
			\hline
			Model & Method & \tabincell{c}{Top-1 \\ Err. (\%)} & \tabincell{c}{Top-5 \\ Err. (\%)} & Time (hour) \\ 
			\hline\hline
			\multirow{2}{*}{{\tabincell{c}{ResNet-56 \\ (CIFAR-10)}}} & DCP \cite{zhuang2018discrimination} & 6.51 & - & 7.58\\
			& DCP & \textbf{6.26} & - & \textbf{2.83} \\
			\hline
			\multirow{2}{*}{{\tabincell{c}{ResNet-18 \\ (ILSVRC-12)}}} & DCP \cite{zhuang2018discrimination} & \textbf{35.88}& \textbf{14.32} & 31.78\\
			& DCP & 35.89 & 14.34 & \textbf{5.18} \\
			\hline
		\end{tabular}
	}
\end{table}

\subsection{Effect of the improved training techniques} \label{sec:effective_training}

\subsubsection{Effect of the single round fine-tuning}
\label{sec:effect_single_shot_finetuning}
To evaluate the effectiveness of the single round fine-tuning, we prune 50\% channels of ResNet-56 on CIFAR-10 and 70\% channels of ResNet-18 on ILSVRC-12. We compare the proposed DCP with the conference version \cite{zhuang2018discrimination} and report the testing error and time of channel pruning in Table~\ref{table:comparison_NIPS_TPAMI}. From the table, the proposed DCP with single round fine-tuning significantly reduces the training time while maintaining comparable performance to the previous DCP~\cite{zhuang2018discrimination}. \liu{For example, DCP pruned ResNet-18 on ILSVRC-12 reduces the computational cost by \textbf{6.14$\times$} compared with the conference version.} These results demonstrate the effectiveness and efficiency of the improved fine-tuning strategy.

\begin{table}[!t]
	\renewcommand{\arraystretch}{1.3}
	\caption{Effect of feature reusing. We report the testing error (\%) and the time of channel pruning on CIFAR-10 and ILSVRC-12.}
	\label{table:explore_feature_reusing}
	\centering
	\scalebox{0.88}{
		\begin{tabular}{c|c||cc|c}
			\hline
			Model & Method & \tabincell{c}{Top-1 \\ Err. (\%)} & \tabincell{c}{Top-5 \\ Err. (\%)} & Time (hour) \\
			\hline\hline
			\liu{\multirow{2}{*}{{\tabincell{c}{ResNet-56 \\ (CIFAR-10)}}}} & without feature reusing & \textbf{6.26} & - & 2.83 \\
			& with feature reusing & 6.28 & - & \textbf{1.85} \\
			\hline
			\liu{\multirow{2}{*}{{\tabincell{c}{ResNet-18 \\ (ILSVRC-12)}}}} &  without feature reusing & \textbf{35.89} & \textbf{14.34} & 5.18 \\
			 &  with feature reusing & 35.90 & \textbf{14.34} & \textbf{3.02} \\
			\hline
	    \end{tabular}
	}
\end{table}

\begin{table}[!t]
	\renewcommand{\arraystretch}{1.3}
	\caption{Pruning results \jing{of} ResNet-56 with different $B$. We report the testing error (\%) and the time of channel pruning on CIFAR-10.}
	\label{table:explore_B_resnet56}
	\centering
	\scalebox{0.88}{
	\begin{tabular}{l||c|c}
		\hline
		$B$ & Testing Err. (\%) & Time (hour) \\
		\hline\hline
		1 & \textbf{6.22} & 1.85 \\
		2 & 6.28 & 0.90 \\
		4 & 6.32 & \textbf{0.48} \\
		\hline
	\end{tabular}
	}
\end{table}

\subsubsection{Effect of the feature reusing}
\label{sec:effect_feature_reusing}
To study the effect of the feature reusing, \liu{we prune 50\% channels of ResNet-56 on CIFAR-10 and 70\% channels of ResNet-18 on ILSVRC-12.}
We report the testing error and time of channel pruning in Table~\ref{table:explore_feature_reusing}. As shown in the table, pruning with the feature reusing achieves comparable performance to the model without the feature reusing. However, pruning with the feature reusing greatly reduces the time of channel pruning. \liu{For example, DCP pruned ResNet-18 with the feature reusing reduces the computational cost by \textbf{1.72}$\times$.} 
Due to the superior performance of the feature reusing, we use it by default in our experiments.

\subsection{Effect of the hyperparameter $B$}
\label{sec:effect_B}
To evaluate the effect of $B$, we prune 50\% channels of ResNet-56 on CIFAR-10 with different $B$ and report the results in Table~\ref{table:explore_B_resnet56}. 
Here, a larger $B$ indicates that we select more channels at each iteration in Algorithm~\ref{alg:channel_selection}. As a result, fewer iterations are required.
From Table~\ref{table:explore_B_resnet56}, the channel pruning time decreases with the increase of $B$ while the performance of DCP degrades. \liu{For example, DCP pruned ResNet-56 with $B=4$ reduces the computational cost by 4.85$\times$}. To make a trade-off between accuracy and complexity, we set $B$ to 2 in our experiments.

\begin{table}[!t]
	\renewcommand{\arraystretch}{1.3}
	\caption{Effect of $\epsilon$ for channel selection. We prune VGGNet with different $\epsilon$ and report the testing error (\%) on CIFAR-10.}
	\label{table:explore_eps}
	\centering
	\scalebox{0.88}{
		\begin{tabular}{cc||c|cc}
			\hline
			\tabincell{c}{Model} & $\epsilon$ & Top-1 Err. (\%) & \#Param. $\downarrow$ (\%) & \#FLOPs $\downarrow$ (\%) \\
			\hline\hline
			\multirow{3}{*}{\tabincell{c}{with \textbf{stopping}\\ \textbf{condition 1}}}
			& $5 \times 10^{-4}$ & 6.29& \textbf{96.16}& \textbf{67.87}\\
			& $3 \times 10^{-4}$ & 5.91& 94.59& 56.73\\
			& $1 \times 10^{-4}$ & \textbf{5.31}& 90.93& 30.80\\
			\hline
			\multirow{3}{*}{\tabincell{c}{with \textbf{stopping}\\ \textbf{condition 2}}}
			& $5 \times 10^{-4}$ & 6.36& \textbf{95.95}& \textbf{84.61}\\
			& $3 \times 10^{-4}$ & 5.87& 94.74& 79.66\\
			& $1 \times 10^{-4}$ & \textbf{5.45}& 91.69& 69.81\\
			\hline
	\end{tabular}}
\end{table}

\subsection{Effect of the tolerance $\epsilon$ in stopping conditions}
\label{sec:effect_eps}
We test different tolerance values in Eq. (\ref{eq:stop_con_outer}) and Eq. (\ref{eq:stopping_condition_2}). Here, we prune VGGNet on CIFAR-10 with $\epsilon\in\{1 \times 10^{-4}, 3 \times 10^{-4}, 5 \times 10^{-4}\}$. We show the experimental results in Table~\ref{table:explore_eps}. In general, a smaller $\epsilon$ will lead to a more rigorous stopping condition. Hence, more channels will be selected. As a result, the performance of the pruned models is improved with the decrease of $\epsilon$. This experiment demonstrates the effectiveness of stopping conditions for automatically determining the pruning rate.
\begin{table}[!t]
	\renewcommand{\arraystretch}{1.3}
	\caption{Pruning results \jing{of} ResNet-56 with different stopping conditions. We report the testing error (\%) on CIFAR-10.}
	\label{table:effect_upper_bound}
	\centering
	\scalebox{0.88}{
	\begin{tabular}{c||c|cc}
		\hline
		Model & \tabincell{c}{Testing \\Err. (\%)} & \tabincell{c}{\#Param.  $\downarrow$ (\%)} & \tabincell{c}{\#FLOPs $\downarrow$ (\%)} \\
		\hline\hline
		with \textbf{stopping condition 1} & 6.36 & 65.20 & 46.82 \\
		with \textbf{stopping condition 2} & \textbf{6.23} & \textbf{68.48} & \textbf{54.80} \\
		\hline
	\end{tabular}
	}
\end{table}
\vspace{-0.1in}


\begin{figure*}[!t]
	\centering
	\subfigure[FLOPs \wrt each layer in the pruned ResNet-56.]
	{
	    \label{fig:effect_upper_bound_vgg_channels}
	    \includegraphics[width=0.46\linewidth]{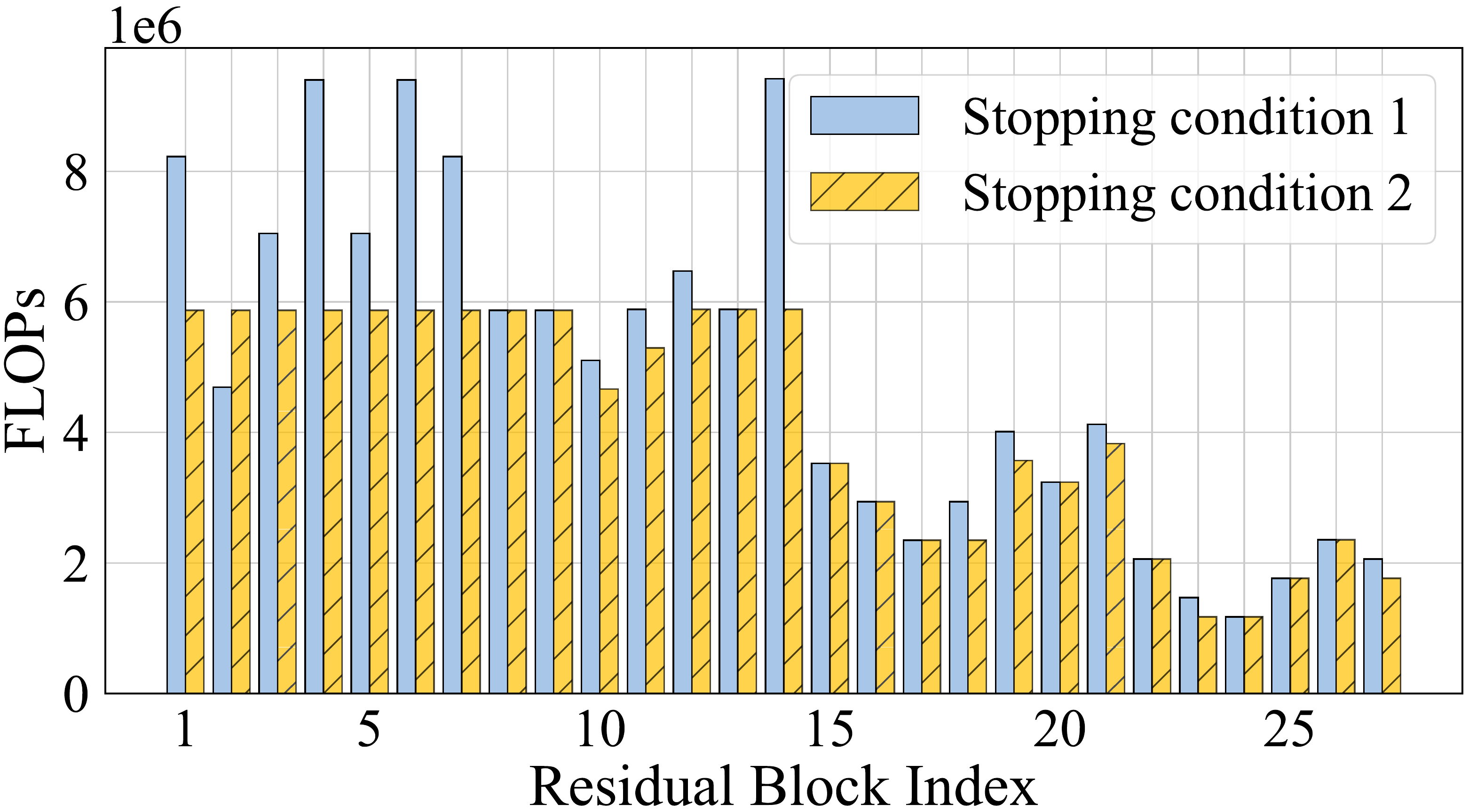}
	}
	\subfigure[\liu{FLOPs w.r.t. each layer in the pruned VGGNet.}]
	{
	    \label{fig:effect_upper_bound_vgg_flops}
	    \includegraphics[width=0.46\linewidth]{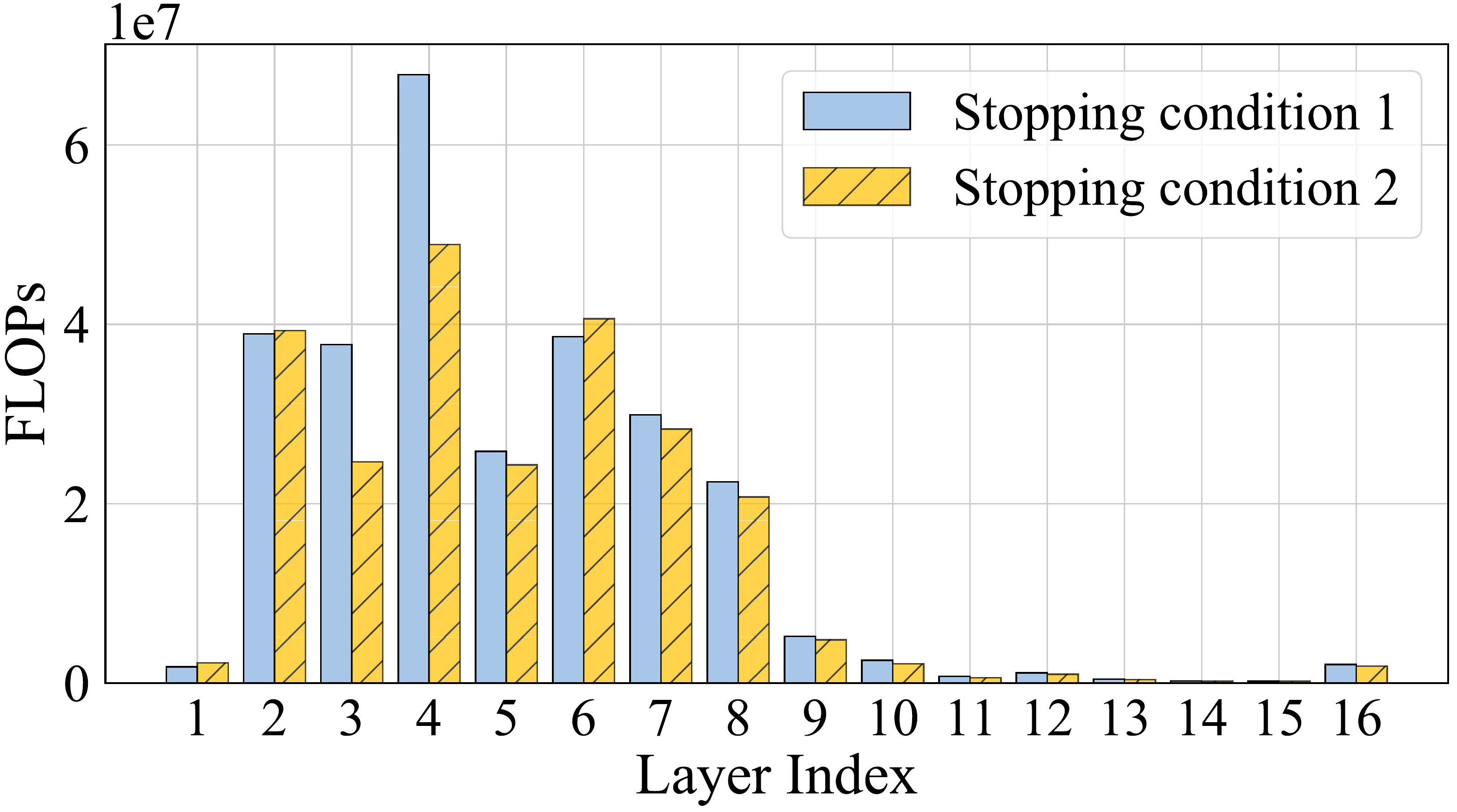}
	}
	\vspace{-0.1in}
	\caption{Number of FLOPs \wrt each layer in the pruned ResNet-56 and VGGNet with different stopping conditions. }
	\label{fig:effect_upper_bound_vgg}
\end{figure*}

\begin{figure*}[!t]
    \centering
	\includegraphics[width=0.85\linewidth]{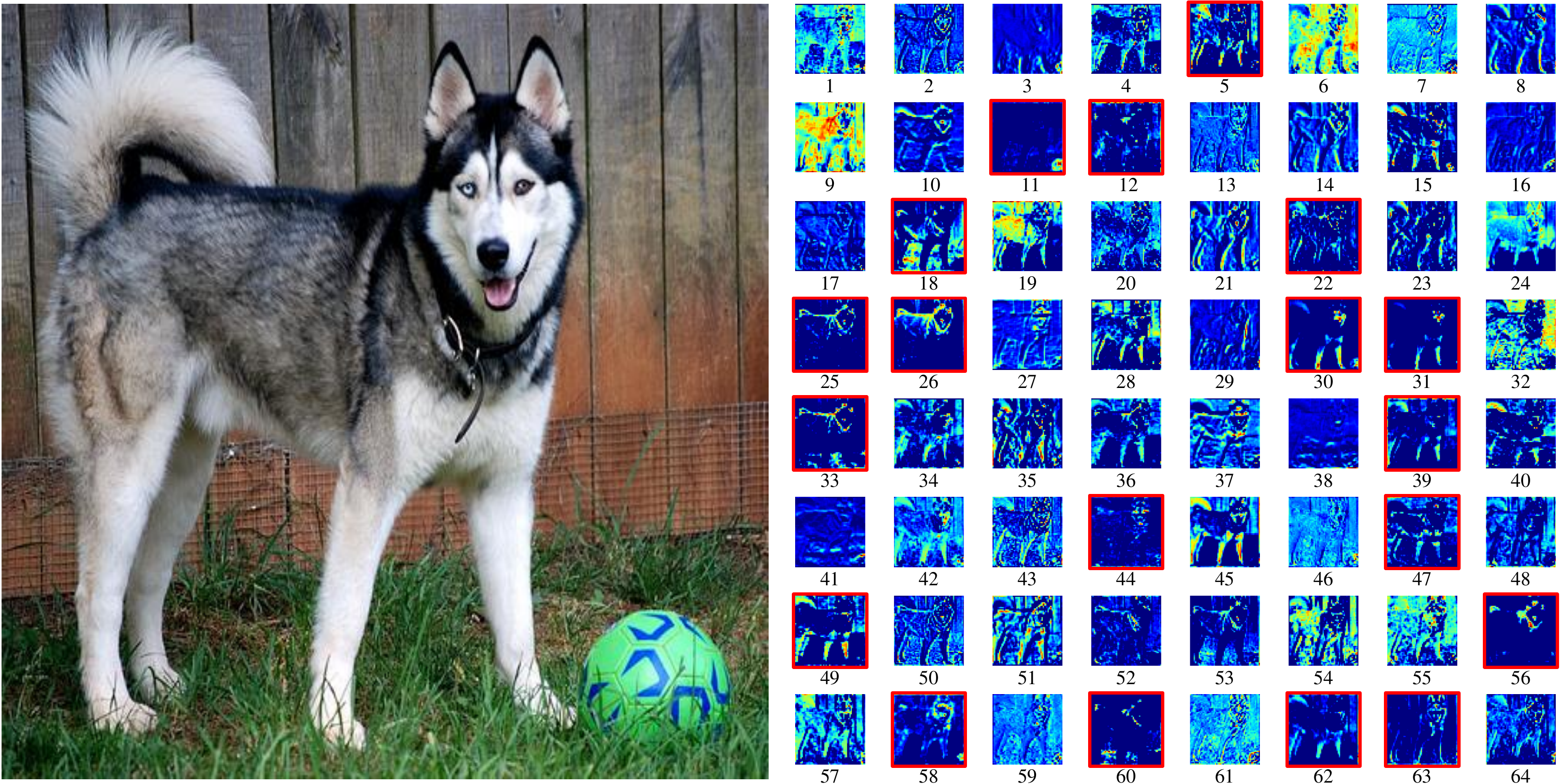}
    \caption{\liu{Visualization of the feature maps from the second layer of the first residual block in ResNet-18 on ILSVRC-12. Feature maps with and without the red bounding boxes are the pruned and selected channels, respectively.}}
    \label{fig:visualization_feature_r18}
\end{figure*}

\subsection{Effect of different stopping conditions}
\label{sec:effect_upper_bound}
We investigate the effect of different stopping conditions mentioned in Section~\ref{sec:stop_conditions}.
To this end, we prune VGGNet and ResNet-56 using Adapt-DCP with different stopping conditions and report the results in \liu{Tables~\ref{table:explore_eps} and}~\ref{table:effect_upper_bound}.
\ice{From the results, the resultant models with \textbf{stopping condition 2} obtain comparable performance but significantly lower computational costs than the ones of \textbf{stopping condition 1}.}
For example, VGGNet pruned with \textbf{stopping condition 2} reduces 69.81\% FLOPs while the one pruned with \textbf{stopping condition 1} only reduces 30.80\% FLOPs when $\epsilon = 1e^{-4}$. 
To further explore the effect of different stopping conditions, we visualize the number of FLOPs \wrt each layer of the pruned ResNet-56 and VGGNet. From Figure~\ref{fig:effect_upper_bound_vgg}, 
\ice{the proposed \textbf{stopping condition 2} is able to prevent DCP from selecting too many channels, and significantly reduces the computational cost of the compressed models.}
\liu{Due to the superior performance of \textbf{stopping condition 2} in FLOPs reduction},
we use it by default. 

\subsection{Visualization of feature maps}
\label{sec:visualize_feature}
We further visualize all the feature maps \wrt the pruned and selected channels from the first residual block in ResNet-18 in Figure~\ref{fig:visualization_feature_r18}.
From the results, the feature maps of the pruned channels are less informative than those of the selected channels. \liu{Moreover, the proposed DCP is able to remove those highly activated channels that are redundant or nearly identical along with the weakly activated channels. For example, DCP removes the 49-th channel since it is similar to the 45-th channel (cosine similarity: 0.82). Even a channel with high activation values (\eg, the 5-th channel), DCP still removes it as it does not contribute to the discriminative power of the network.}
These results prove that the proposed DCP selects the channels with strong discriminative power for the network. 

\begin{table}[!t]
	\renewcommand{\arraystretch}{1.3}
	\caption{
	Effect of the number of additional losses. We prune 50\% channels from ResNet-56 and 30\% channels from MobileNetV1. We report the testing error on CIFAR-10. The testing error (\%) of pre-trained ResNet-56 and MobilenNet v1 are 6.26 and 6.04, respectively.}
	\label{table:explore_num_losses}
	\centering
    \scalebox{0.85}{
    \begin{tabular}{c||cc}
		\hline
		\multirow{2}{*}{\#Additional Losses} & \multicolumn{2}{c}{Err. gap (\%)} \\
		\cline{2-3}
		& ResNet-56 & MobileNetV1 \\
		\hline\hline
		1 & +0.10 & -0.28 \\
		3 & +0.02 & -0.54 \\
		5 & +0.01 & -0.65 \\
		7 & -0.08 & -0.70 \\
		\hline
	\end{tabular}
    }
\end{table}


\subsection{\guo{Effect of} the number of additional losses}
\label{sec:explore_num_losses}
We prune 50\% channels from ResNet-56 \jing{and 30\% channels from MobileNetV1 with different number of additional losses on CIFAR-10}.
From Table~\ref{table:explore_num_losses}, adding more losses results in better performance. For example, ResNet-56 with three additional losses outperforms the one with one additional loss. However, adding too many losses may lead to a little gain in performance but significantly increase the computational cost. Heuristically, we find that adding losses every 5-10 blocks is sufficient to make a good trade-off between accuracy and complexity.


\section{Conclusion}
In this paper, we have proposed a discrimination-aware channel pruning (DCP) method for the compression of deep neural networks. \liu{Specifically, we first introduce additional losses to improve the discriminative power of the network and then}
perform channel selection in a layer-wise manner. In order to select informative channels, we have formulated the channel pruning as a sparsity-induced optimization problem and proposed a greedy algorithm to solve it. Based on DCP, we have further proposed several techniques to accelerate the channel pruning process. 
Moreover, we have also proposed a discrimination-aware kernel pruning (DKP) method to perform model compression \liu{at} the kernel level. Extensive results on both image classification and face recognition tasks have demonstrated the effectiveness of the proposed methods.

\ice{In the future, we may extend our method in two aspects. First, in the proposed DCP, we perform channel/kernel selection in a layer-wise manner. This strategy, however, may result in suboptimal performance as only a single layer is considered each time. To address this, we may consider multiple layers/blocks each time to improve the pruning performance. Second, we will extend the proposed DCP scheme for model quantization. Specifically, based on DCP, we can conduct the quantization for each selected channel/kernel, and use the reconstruction residual
to choose the next channel/kernel. In this way, we can perform channel/kernel pruning and quantization simultaneously.}

\ifCLASSOPTIONcompsoc
  \section*{Acknowledgments}
\else
  \section*{Acknowledgment}
\fi
This work was partially supported by Key-Area Research and Development Program of Guangdong Province 2018B010107001, National Natural Science Foundation of China (NSFC) 61836003 (key project), National Natural Science Foundation of China (NSFC) 62072190, and Program for Guangdong Introducing Innovative and Enterpreneurial Teams 2017ZT07X183.

\ifCLASSOPTIONcaptionsoff
  \newpage
\fi
\bibliographystyle{abbrv}
{
	\bibliography{reference}
}
\end{document}